\documentclass[10pt,journal,compsoc]{IEEEtran}

\usepackage{graphicx}
\usepackage{amsmath}
\usepackage{amsfonts}
\usepackage{multirow}
\usepackage{siunitx}
\usepackage{booktabs}
\usepackage{tipa}
\usepackage{algorithm}
\usepackage[noend]{algpseudocode}
\usepackage{multirow}
\usepackage{booktabs}

\newcommand{\yh}[1]{{\color{black}#1}}
\newcommand{\yl}[1]{{\color{black}#1}}
\newcommand{\yln}[1]{{\color{black}#1}}
\newcommand{\hyh}[1]{{\color{black}#1}} 

\ifCLASSOPTIONcompsoc
  \usepackage[nocompress]{cite}
\else
  \usepackage{cite}
\fi

\begin{document}

\title{NeRF-Texture: \yln{Synthesizing Neural Radiance Field Textures}}

\author{Yi-Hua~Huang,
        Yan-Pei~Cao,
        Yu-Kun~Lai,
        Ying~Shan
        and~Lin~Gao\IEEEauthorrefmark{1}
        \IEEEcompsocitemizethanks{\IEEEcompsocthanksitem \IEEEauthorrefmark{1} Corresponding Author is Lin Gao (gaolin@ict.ac.cn).
\IEEEcompsocthanksitem Yi-Hua Huang and Lin Gao are with the Beijing Key Laboratory of Mobile Computing and Pervasive Device, Institute of Computing Technology, Chinese Academy of Sciences. Yi-Hua Huang and Lin Gao are also with University of Chinese Academy of Sciences, Beijing, China.
\protect\\
E-mail: \{huangyihua20g, gaolin\}@ict.ac.cn
\IEEEcompsocthanksitem Yan-Pei~Cao is with VAST.
\protect\\
E-mail: caoyanpei@gmail.com
\IEEEcompsocthanksitem Ying~Shan is with ARC Lab, Tencent PCG, China.
\protect\\
E-mail: yingsshan@tencent.com
\IEEEcompsocthanksitem Yu-Kun Lai is with School of Computer Science \& Informatics, Cardiff University, UK.
\protect\\
E-mail: LaiY4@cardiff.ac.uk
}}

\markboth{Accepted by IEEE Transactions on Pattern Analysis and Machine Intelligence}%
{Shell \MakeLowercase{\textit{et al.}}: Bare Demo of IEEEtran.cls for Computer Society Journals}

\IEEEtitleabstractindextext{%
\begin{abstract}
Texture synthesis is a fundamental problem in computer graphics that would benefit various applications. Existing methods are effective in handling 2D image textures. In contrast, many real-world textures contain meso-structure in the 3D geometry space, such as grass, leaves, and fabrics, which cannot be effectively modeled using only 2D image textures. We propose a novel texture synthesis method with Neural Radiance Fields (NeRF) to capture and synthesize textures from given multi-view images. In the proposed NeRF texture representation, a scene with fine geometric details is disentangled into the meso-structure textures and the underlying base shape. This allows textures with meso-structure to be effectively learned as latent features situated on the base shape, which are fed into a NeRF decoder trained simultaneously to represent the rich view-dependent appearance. Using this implicit representation, we can synthesize NeRF-based textures through patch matching of latent features. However, inconsistencies between the metrics of the reconstructed content space and the latent feature space may compromise the synthesis quality. To enhance matching performance, we further regularize the distribution of latent features by incorporating a clustering constraint. 
\yln{In addition to generating NeRF textures over a planar domain, our method can also synthesize NeRF textures over curved surfaces, which are practically useful.}
Experimental results and evaluations demonstrate the effectiveness of our approach.
\end{abstract}

}

\maketitle

\IEEEdisplaynontitleabstractindextext

\IEEEpeerreviewmaketitle

\begin{figure*}
  \includegraphics[width=\textwidth]{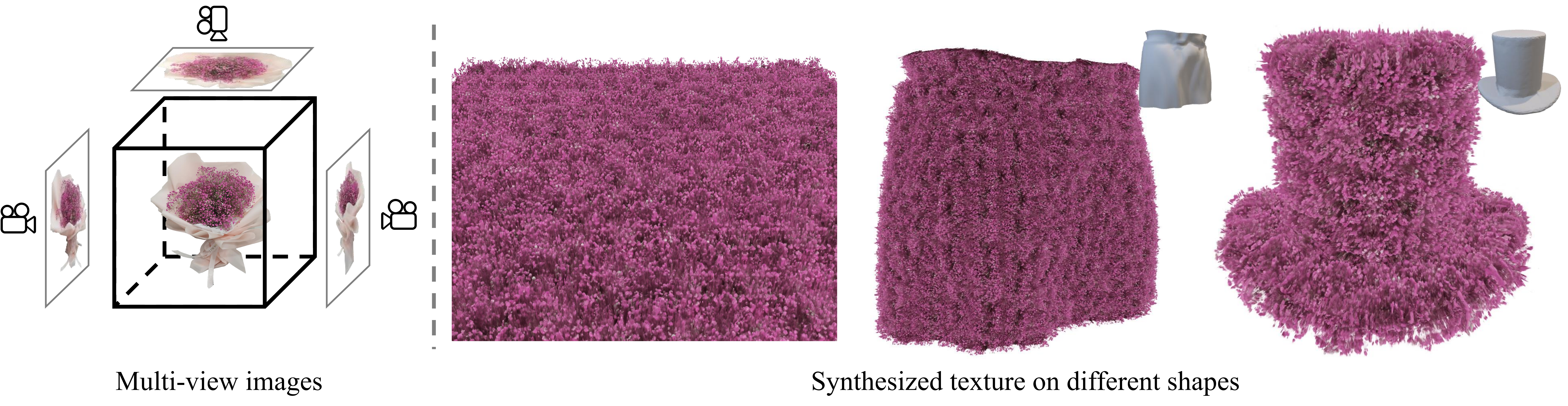}
  \vspace{-6mm}
  \caption{Given a set of multi-view images of the target texture with meso-structure, our model synthesizes Neural Radiance Field (NeRF) textures, which can then be applied to novel shapes, such as the skirt and hat in the figure, with rich geometric and appearance details.}
  \label{fig:teaser}
\end{figure*}



\IEEEraisesectionheading{\section{Introduction}\label{sec:introduction}}
\IEEEPARstart{C}{apturing}, modeling, synthesizing, and rendering real-world textures are fundamental problems in computer graphics and computer vision.
{In the real world}, textures with high-frequency geometry are ubiquitous, like grass, leaves, fabrics, and cobblestones. Unfortunately, it is intractable to directly model such meso-structure with polygons, curves, or voxels~\cite{baatz2021nerf}, like flowers shown in Fig.~\ref{fig:teaser}.
While a conventional texture map can represent a range of surface properties, such as color, reflection, transparency, and displacement, it remains impractical to accurately portray view-dependent appearance and meso-structure~\cite{kuznetsov2022rendering}.

Thanks to recently proposed neural implicit rendering approaches such as NeRF (Neural Radiance Fields)~\cite{mildenhall2021nerf}, textures in complex real scenes could be reconstructed from multi-view images.
The vanilla NeRF mixes the representation of geometry and appearance, which limits the freedom to 
manipulate the reconstructed textures.
To support texture swapping and editing, NeuMesh~\cite{yang2022neumesh} and NeuTex~\cite{xiang2021neutex} make an attempt to disentangle the texture and geometry.
NeuMesh~\cite{yang2022neumesh} supports geometry and appearance editing but is incapable of modeling and synthesizing meso-structure textures; NeuTex~\cite{xiang2021neutex} parameterizes the scene content in 3D Euclidean space over 2D UV space, which is suitable for modeling smooth surfaces rather than high-frequency meso-structure textures.

In computer graphics applications, once texture samples are captured, texture synthesis is an essential step to produce similar (but not repetitive) larger textures to decorate a target surface. Although there is extensive research on 2D image texture synthesis, little attention has been paid to the synthesis of NeRF-based textures.

In this paper, we propose a novel NeRF-based approach for capturing, modeling, synthesizing, and applying textures with meso-structure and view-dependent appearance, leveraging multi-view images obtained from real-world scenes.
Our method only requires a set of multi-view images of the texture to acquire as input, which can be easily obtained by shooting a short video using a mobile phone. 
Our approach then learns the representations of the texture and synthesizes it to the desired size over a UV parameter space, typically in several minutes.
Ultimately, the synthesized NeRF texture can be applied to any given shape, enabling real-time rendering.

More specifically, we propose the following key techniques to facilitate the modeling and synthesis of the NeRF textures with detailed geometry and view-dependent effects:

Firstly, to learn the meso-structure of textures, we disentangle the scene with fine geometric details into the meso-structure and the underlying base shape. We then learn the meso-structure as a NeRF texture through a latent feature field defined on the base shape. To achieve this goal, we first extract the base shape and explicitly represent it as a coarse mesh using Instant-NGP~\cite{mueller2022instant} and Co-ACD~\cite{wei2022approximate}. We then propose to map each point in the 3D Euclidean space to the Cartesian product of the signed distance and its foot point when projected onto the base mesh. Latent features are defined on the base shape and fetched by the foot point.
However, directly fetching latent codes from mesh vertices, like in~\cite{yang2022neumesh}, requires high-resolution meshes, which leads to a slowdown in latent code lookup and requires distillation from a well-trained NeRF.
Instead, we fetch the latent representation for the texture with hash grids~\cite{mueller2022instant} to support real-time rendering and training from scratch. 

Secondly, to apply captured NeRF-based textures to new shapes, it is crucial to synthesize textures at sufficient resolutions. We propose a novel NeRF-based texture synthesis method based on the coarse-fine disentanglement representation. Initially, we extract implicit patches from the base shape, on which latent features are defined, to create a patch collection. Subsequently, we implement an implicit patch matching algorithm to synthesize NeRF-based textures with collected patches. During this process, patches of latent features are sampled, matched and quilted to generate a texture space with the desired spatial resolution. Furthermore, we introduce an unsupervised metric learning approach to cluster the features of similar textures, thereby enhancing the quality of the synthesized results.
\yln{In this paper, we significantly extended our previous conference paper~\cite{huang2023nerf-texture} by extending the method to synthesize NeRF textures over curved surfaces as well as including more extensive evaluation, including results on newly captured datasets and more ablation study.}

In summary, our main contributions are as follows:
\begin{itemize}
    \item We propose a method to capture, model, synthesize and render NeRF textures with meso-structure from real-world multi-view images.
    \item We propose a coarse-fine disentanglement representation that learns the meso-structure and reflection coefficients as NeRF textures, which are separated from the underlying coarse surface.
    \item We adopt a patch matching algorithm in the latent space to synthesize NeRF textures. A clustering constraint is introduced to regularize the latent distribution for better matching. \yl{The method is further extended to NeRF texture synthesis on curved surfaces.} To the best of our knowledge, this is the first work for NeRF texture synthesis.
\end{itemize}

\begin{figure*}[htbp]
	\centering
	\includegraphics[width=1.0\linewidth]{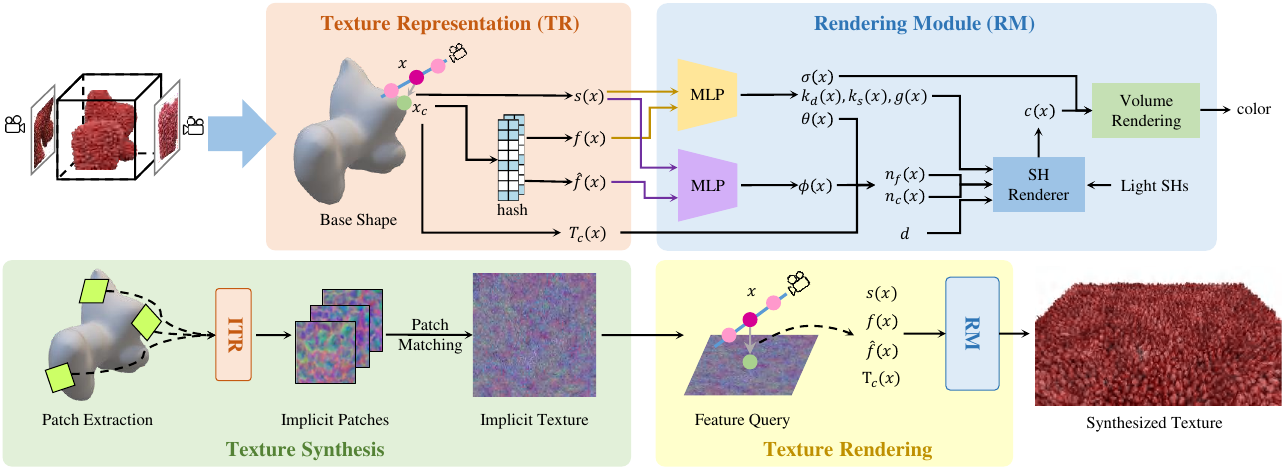}
	\vspace{-5mm}
	\caption{\textbf{Overview of our method.} Given a set of multi-view images, we first estimate its base shape. Based on it, we model the scene with a disentangled representation of the base shape and NeRF texture with meso-structure. The query point $x$ is projected onto the base shape as footpoint $x_c$. Latent features $f(x),\hat{f}(x)$ representing textures are fetched by feeding $x_c$ to hash grids. Along with matrices of local tangent space $T_c(x)$, latent features $f(x),\hat{f}(x)$, and SDF value $s(x)$ are fed into the rendering module (RM). The density $\sigma(x)$, coefficients of Phong shading model $k_d(x),k_s(x),g(x)$, elevation and azimuth angles of the fine normal $\theta(x),\phi(x)$ are predicted based on the input features and SDF.
    The color $c(x)$ of the query point $x$ is calculated by Spherical Harmonic (SH) rendering based on the coarse and fine normals $n_c(x), n_f(x)$, viewing direction $d$, shading coefficients $k_d(x),k_s(x),g(x)$ and lighting SHs.
    Based on the implicit texture representation (ITR), we extract implicit patches from the base shape and synthesize texture by an implicit patch matching algorithm. By querying $f(x),\hat{f}(x)$ and $T_c(x)$ from the synthesized implicit textures, we are able to render the appearance of the synthesized texture.}
\label{fig:pipeline}
\end{figure*}


\section{Related Work}
\label{sec:related}

As our work is related to neural rendering and texture synthesis, we review papers related to these topics.

\subsection{Neural Rendering}
Various neural rendering approaches have been proposed to synthesize novel views of a scene with a given set of photographs. 
NeRF~\cite{mildenhall2021nerf} models the scene as a radiance field with particles emitting and blocking lights. Inspired by NeRF, follow-up works extend it to achieve faster inference~\cite{fridovich2022plenoxels,karnewar2022relu,mueller2022instant}, handle large-scale scenes~\cite{zhang2020nerf++,barron2022mip,tancik2022block,gao2023surfelnerf} and dynamic scenes~\cite{NEURIPS2022_eeb57fdf,qiao2022neuphysics,song2023nerfplayer}, and attain reflection decomposition~\cite{boss2021nerd,srinivasan2021nerv,kuang2022neroic,munkberg2022extracting} and stylization~\cite{zhang2022arf,huang2022stylizednerf,fan2022unified}.
Some other neural representations have been proposed to model meso-scale textures.
Kuznetsov et al.~\cite{kuznetsov2022rendering} utilize neural bidirectional texture functions (BTFs) to model known texture with meso-structure.
Wang et al.~\cite{yifan2021geometry} propose to learn a complex shape as a combination of a smooth low-frequency signed distance function (SDF) and a continuous high-frequency signed distance function.
\yh{Concurrent work~\cite{li2023nerf} synthesizes DVGO \yln{(Direct Voxel Grid Optimization)}~\cite{sun2022direct}-based 3D scenes with the guidance of shading maps.}

NeuTex~\cite{xiang2021neutex} explicitly represents the texture in a neural representation through UV parameterization to support texture editing and mapping. 
However, such 2D parameterization assumes the target object can be smoothly mapped to a 2D parameter space, which is not suitable for most textures with meso-structure.
NeuMesh~\cite{yang2022neumesh} proposed a mesh-based neural implicit representation to disentangle the shape and appearance. 
With geometry and texture features defined on vertices, it achieves the geometry and texture editing of the neural implicit field.
Nevertheless, NeuMesh utilizes predicted SDF rather than densities in volume rendering, which cannot be defined on non-watertight meso-structure. Besides, the mesh storing encodings closely fits the target surface, and as a result the meso-structure is not learned as texture properties.
NeRF-Tex~\cite{baatz2021nerf} firstly investigated the possibility to model the texture with meso-structures through NeRF. 
The model is trained on synthetic datasets with rendering results of patches in a bounding box on a plane under known lighting conditions.
Textures are mapped to the shapes by repeatedly placing the reconstructed bounding box on surfaces.
In contrast, our approach targets NeRF texture synthesis, which simultaneously learns the Phong reflection coefficients, meso-structure and lighting conditions from real-world objects with textures.

\subsection{Texture Synthesis}
The goal of texture synthesis is to synthesize a new texture that appears to be generated by the same underlying process~\cite{wei2009state}. The pioneering work by \cite{efros1999texture} gradually grows the synthesized region by assigning pixels one by one. The assignment is determined by neighborhood similarity. 
Following this idea, a fixed neighborhood is used in~\cite{wei2000fast} to avoid non-uniform pattern distribution. Patch-based method~\cite{liang2001real} proposes to blend the overlapped regions between patches. The works~\cite{efros2001image,kwatra2003graphcut} cut through the overlapped regions via dynamic programming and graph cut, respectively. PatchNet~\cite{hu2013patchnet} searches an image library to locate ideal regions adhering to the synthesis constraints.
Kwatra et al.~\cite{kwatra2005texture} proposed an alternative approach by texture optimization.

In addition to traditional matching and optimization methods, neural networks are also introduced in texture synthesis. Gatys et al.~\cite{gatys2015texture} present a data-driven approach to generating texture through optimizing the Gram matrix of latent features extracted by VGG network~\cite{simonyan2014very}.
Follow-up works \cite{ulyanov2016texture,johnson2016perceptual} train feed-forward convolutional networks to replace the time-consuming optimization process. 
Generative adversarial networks (GANs) are also widely used for texture synthesis~\cite{jetchev2016texture,li2016precomputed}.
Zhou et al.~\cite{zhou2018non} train a GAN to double the spatial extent of texture blocks, enabling the model to synthesize non-stationary texture.
Portenier et al.~\cite{portenier2020gramgan} use the Gram matrix produced by the discriminator in adversarial loss to improve the quality of synthesized texture.
Hertz et al.~\cite{Hertz2020deep} propose a Mesh-CNN~\cite{hanocka2019meshcnn} based GAN architecture to synthesize geometric textures.
PSGAN~\cite{bergmann2017learning} proves that periodic encoding can improve the quality of GAN results.
Inspired by it, Chen et al.~\cite{chen2022exemplar} utilize periodic embedding as input and replace the convolution layer with a Multi-Layer Perceptron (MLP) to model implicit fields.


\section{Method}
\label{sec:method}

We present a method to capture, model, synthesize and apply NeRF textures with meso-structure from real-world multi-view images.
The overview of our pipeline is shown in Fig.~\ref{fig:pipeline}.
Given segmented multi-view images of the scene, our model learns to disentangle meso-structure textures and the underlying base shape. By sampling the implicit patches of latent features on the base shape and utilizing them to synthesize a larger texture map, we are able to decorate an arbitrary given mesh with the synthesized result.
In the following, we will introduce texture representation in Sec.~\ref{subsec:itr}, texture synthesis in Sec.~\ref{subsec:its}, and model optimization in Sec.~\ref{subsec:optimization}.

\subsection{Texture Representation}
\label{subsec:itr}

\subsubsection{Base Shape Extraction} 
\begin{figure}[htbp]
	\centering
	\includegraphics[width=.95\linewidth]{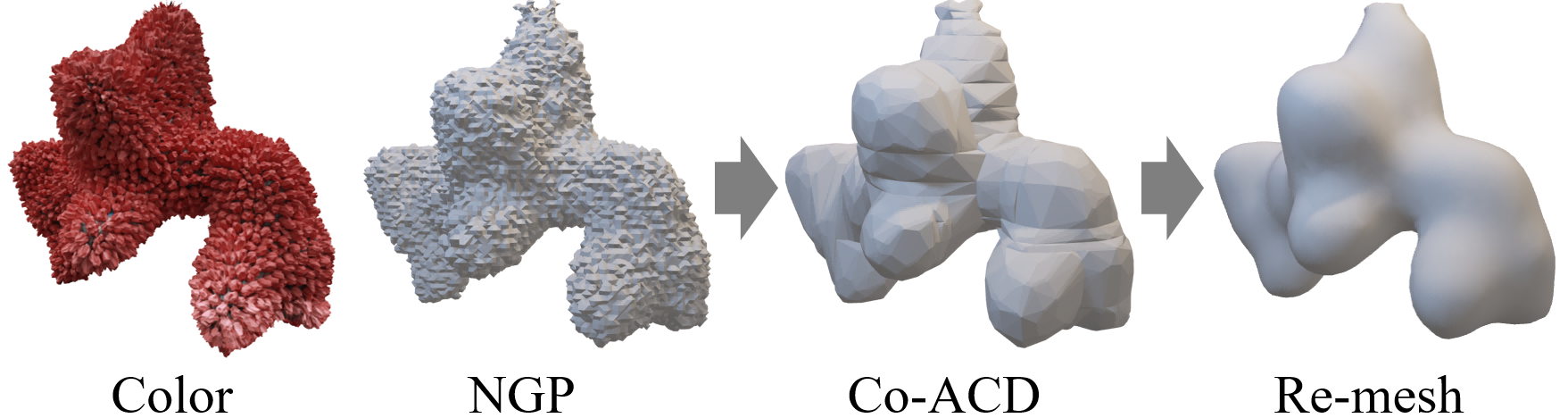}
	\vspace{-2mm}
	\caption{\textbf{Base Shape Extraction.} We show the intermediate outputs during the base shape extraction, including NGP~\cite{mueller2022instant}, Co-ACD~\cite{wei2022approximate}, and re-meshing~\cite{huang2018robust,vollmer1999improved}.}
\label{fig:coarse_shape_obtaining}
\end{figure}

To model the base shape explicitly as a coarse mesh, we firstly adopt Instant-NGP~\cite{mueller2022instant} to reconstruct the coarse mesh by executing Marching Cubes~\cite{lorensen1987marching} on the estimated density field with camera parameters estimated by COLMAP~\cite{schoenberger2016sfm,schoenberger2016mvs}. To remove the meso-structure and make the coarse mesh smoother, the coarse mesh is transferred into the union of approximately decomposed convex hulls by Co-ACD~\cite{wei2022approximate}.
The shape is then re-meshed~\cite{huang2018robust,vollmer1999improved} to a mesh with vertices uniformly distributed on the surface.
Fig.~\ref{fig:coarse_shape_obtaining} illustrates the process of base shape extraction.

\subsubsection{Base Shape Projection} 
\begin{figure}[htbp]
	\centering
	\includegraphics[width=0.9\linewidth]{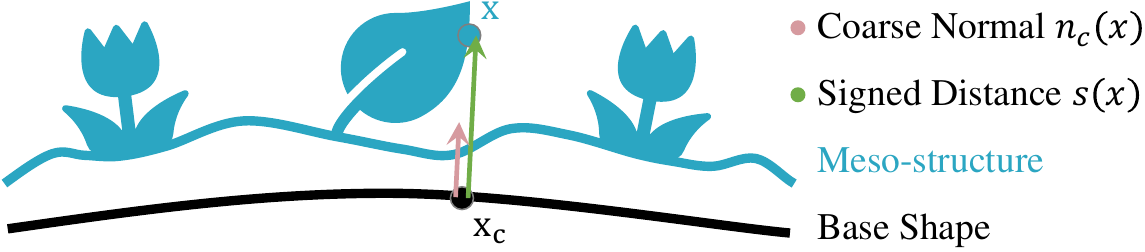}
	\vspace{-3mm}
	\caption{\textbf{Illustration of Base Shape Projection in 2D}. Point 
$x$ in Euclidean space is parameterized as the signed distance $s(x)$ and the projected footpoint $x_c$.}
\label{fig:coarse_fine_disentanglement}
\end{figure}

We treat all attributes other than the base shape as texture attributes to learn, including meso-structure, normal and appearance.
To disentangle these attributes and the base shape, we utilize the coarse mesh mentioned above to re-parameterize 3D Euclidean space and learn the attributes into the latent features defined on the coarse mesh. 
In our approach, the coordinates of query point $x$ are mapped to the coarse mesh to get the projected coordinates $x_c$ and the signed distance $s(x)$, as depicted in Fig.~\ref{fig:coarse_fine_disentanglement}. 
For each query point $x$, we look up its $K(=8)$ nearest neighbor points $\{v_k\}$ among the coarse mesh vertices.
We interpolate the vertex normal $n_v$ of neighbors together with the normalized vector from nearest neighbor $v_1$ to $x$ using weighted KNN~\cite{shepard1968two} to get the coarse mesh normal $n_c$ of $x$:
\begin{equation}
\begin{aligned}
    \tilde{n}_c(x) = \sum\limits_{k=1}^K \frac{1}{W}\left(\frac{n_v(v_k)}{||x-v_k||_2} + \frac{x-v_1}{w||x-v_1||_2}\right), \\
    n_c=\frac{\tilde{n}_c}{||\tilde{n}_c||_2},\ W=\sum\limits_{k=1}^K \frac{1}{||x-v_k||_2}+\frac{1}{w},
\end{aligned}
\label{eq:coarse}
\end{equation}
where $w$ is a constant set to 0.01. Next, we cast the ray from $x$ along the opposite coarse normal direction $-n_c(x)$ to hit the coarse mesh on a projected point $x_c$. The first term in Eq.~\ref{eq:coarse} is the weighted average of normals from the $K$ nearest neighbors to improve robustness. When $x$ is far from the coarse mesh, normals of $K$ neighbors may be less reliable so the second term becomes dominant to ensure the ray-mesh collision.
At this step, the signed distance of $x$ projected onto the coarse mesh is also obtained and denoted as $s(x)$. 

\subsubsection{Differentiable Projection Layer} The step of ray casting makes the projected coordinates $x_c$ non-differentiable with respect to the input coordinates $x$. However, the gradient is essential to approximate the normal~\cite{srinivasan2021nerv,boss2021nerd} or supervise the normal estimation~\cite{kuang2022neroic,zhang2021nerfactor} for physically based rendering.
In addition, back-propagating gradients to the camera parameters via coordinates $x$ is crucial for camera pose modification~\cite{lin2021barf,wang2021nerfmm,kuang2022neroic} to improve the reconstruction quality.
For these reasons, we construct a differentiable projection layer by specifying the following derivation rule:
\begin{equation}
    \frac{\mathrm{d} x_c}{\mathrm{d} x} = (I - n_c(x)^Tn_c(x)),\ \frac{\mathrm{d} s(x)}{\mathrm{d} x} = n_c(x)
\end{equation}%
\noindent where $I$ is the identity matrix. The rule transfers the component of the gradient of $x_c$ on the plane, which is perpendicular to $n_c(x)$, to $x$.
It also passes the gradient of $s(x)$ to $x$ after projection onto $n_c(x)$.
The rule is consistent with parameterizing 3D coordinates as the footpoint and projected signed distance on a base shape.

\subsubsection{Attributes Prediction} 
Directly querying latent codes from mesh vertices, like in~\cite{yang2022neumesh}, is difficult to train from scratch and demands high-resolution meshes, which results in high query overhead and difficulty in real-time rendering.
Hence, we fetch the latent texture representation $f(x)$ in $O(1)$ time complexity by feeding the projected coordinates $x_c$ to hash grids storing latent features~\cite{mueller2022instant}.
Through the tiny-cuda-nn framework~\cite{tiny-cuda-nn}, we map the concatenated texture feature $f(x)$ and Fourier embedded~\cite{tancik2020fourier} SDF value $s(x)$ to the density $\sigma(x)$ and reflection coefficients.
The estimation of the fine normal $n_f(x)$ on $x$ is done in two parts: estimating elevation angle $\theta(x)$ and azimuth angle $\phi(x)$, respectively.
Both angles are represented in the local tangent frame of $x_c$, denoted as $T_c(x_c) = (t(x_c), b(x_c), n(x_c))$, meaning tangent, bitangent, and normal at $x_c$.
Notice that $T_c(x_c)$ is determined by the tangent, bitangent, and normal of $x_c$'s locating triangle face, which is pre-computed and fixed.
Since $\theta(x)$ is the angle between the coarse mesh normal $n_c(x)$ and the fine (meso-structure) normal $n_f(x)$, it is an attribute independent of the definition of the local tangent frame.
Instead, $\phi(x)$ depends on the direction of $t(x_c)$, which can be flexibly pre-chosen.
Hence we predict $\theta$ with $s(x)$ and $f(x)$, which is further used for patch matching, to encourage similar texture contents to have latent features $f$ close to each other regardless of different local tangent definitions.
We then learn a different feature $\hat{f}(x)$ stored in another hash grid table for predicting $\phi(x)$.

\subsubsection{Shading Model}
\begin{figure}[htbp]
	\centering
	\includegraphics[width=1\linewidth]{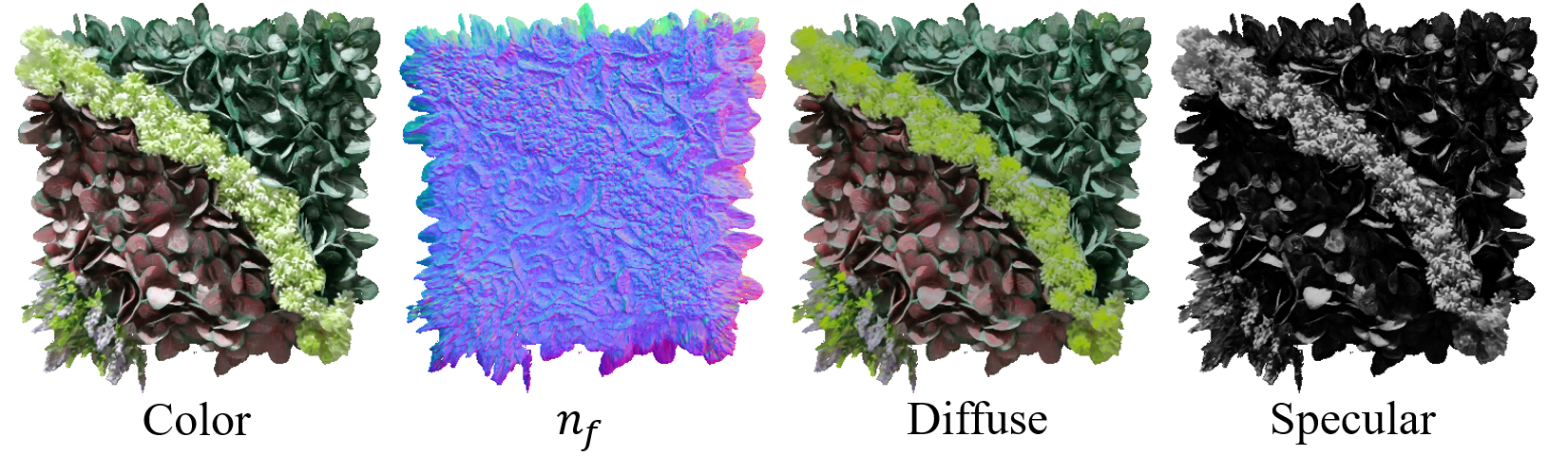}
	\vspace{-7mm}
	\caption{\textbf{Shading Decomposition}. Our model predicts the fine normal $n_f$ and decomposes the radiance into diffuse and specular components.}
\label{fig:lighting_decomposition}
\end{figure}

Unlike vanilla NeRF, which mixes the representation of materials and lighting, we decompose these elements to enable the rendering of textures mapped to novel locations.
To ensure real-time rendering speed and stable convergence, we utilize Spherical Harmonics (SHs)~\cite{ramamoorthi2001signal} to represent illumination and materials in our rendering pipeline.
We adopt Phong shading~\cite{phong1975illumination} to model the material reflection with three parameters: diffuse coefficient $k_d$, specular coefficient $k_s$, and glossiness $g$.
Following the approach outlined in~\cite{ramamoorthi2001signal}, we employ the convolution of SHs to compute the texture color $c(x)$. The decomposition is illustrated in Fig.~\ref{fig:lighting_decomposition}.

\subsection{Texture Synthesis}
\label{subsec:its}

\subsubsection{Texture Patch Extraction}
\begin{figure}[htbp]
	\centering
	\includegraphics[width=1\linewidth]{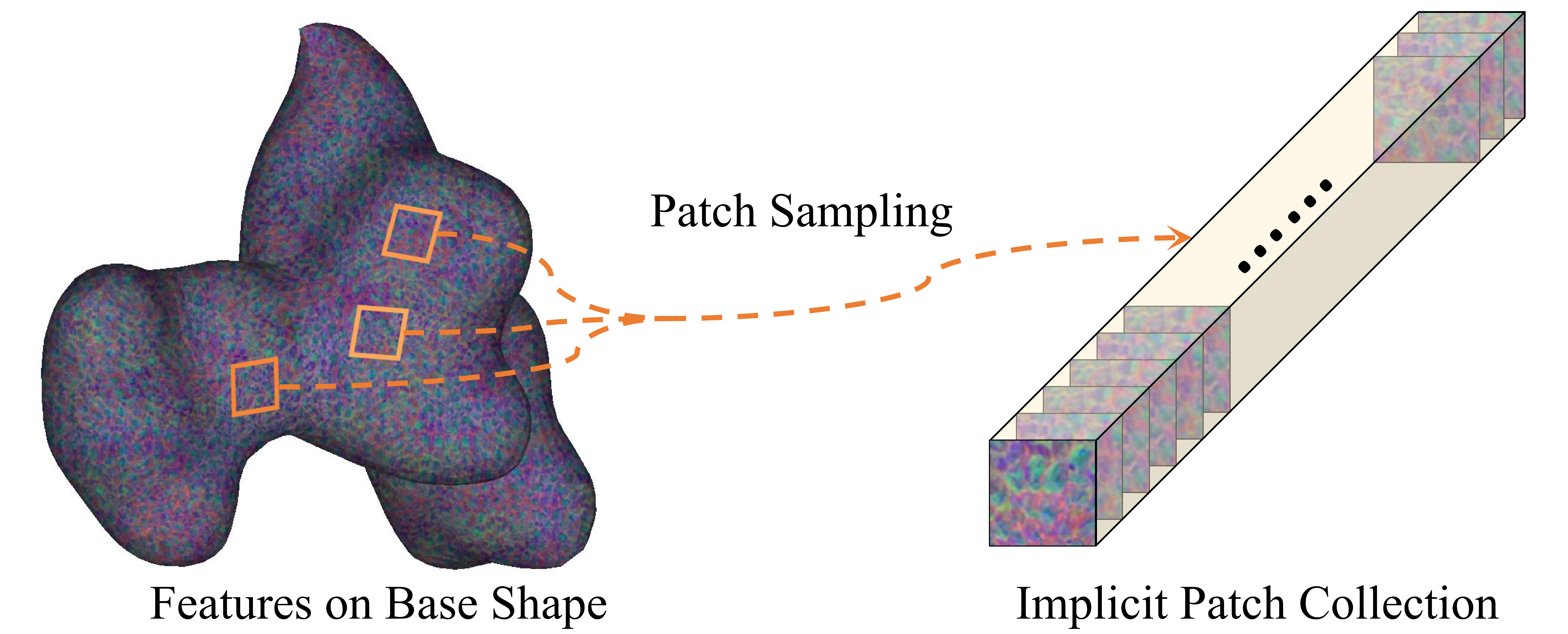}
	\vspace{-6mm}
	\caption{\textbf{Texture Patch Extraction}. We extract implicit texture patches by sampling them on the base shape, where latent features are defined.}
\label{fig:texture_patch_obtaining}
\end{figure}

Since we have leveraged latent features on the base shape for representing texture attributes, the next step is to extract the implicit patches from the base shape for subsequent texture synthesis, as depicted in Fig.~\ref{fig:texture_patch_obtaining}.
\yh{In our approach, we firstly sample centers of patches on the entire base shape or user-specified regions via Poisson disk sampling~\cite{cook1986stochastic} to get evenly distributed points.}
Next, we place square scan arrays of $128 \times 128$ resolution on each tangent plane of the coarse mesh to obtain the intersections of the scanning rays with the mesh.
\yh{We discard patches with too long distance of ray casting to filter out those with excessive curvature.}
We then query the hash grids with these intersections to fetch latent features and obtain implicit patches.
\yh{The obtained patches are denoted as $\{F_i\in \mathbb{R}^{128 \times 128 \times C}\}$, where $C$ is the latent dimension.}
We also denote the rotation of the sampling local frame to the world coordinate system as $T_s\in\mathbb{R}^{3\times3}$.
We similarly define the rotation of the coarse mesh local frame to the world system as $T_c\in\mathbb{R}^{3\times3}$.
For subsequent texture mapping, we also record $T_c$ and $T_s$ of each patch.
The sampled patches are augmented by horizontal and vertical flipping for better synthesis.
The transformation matrices of the sampling tangent space of augmenting patches are also flipped accordingly.

\subsubsection{Patch-based Synthesis}
We synthesize textures of arbitrary sizes based on the sampled exemplars using patch matching and quilting~\cite{efros2001image}.
The output is initialized by copying a seed patch, and the synthesized region is gradually grown from the initial state by iteratively copying the picked patch onto it.
\yh{ The patch is randomly selected from 4 candidates with the most similar overlapping regions. The matching process is accelerated by pre-building a kd-tree of patches' overlapping regions. With a picked patch, the next step is to quilt it with the synthesized texture in the overlapping regions.
Denoting the overlapping region of synthesized output and candidate patches as $F_1^{ov} \in \mathbb{R}^{H \times W \times C}$ and $F_2^{ov} \in \mathbb{R}^{H \times W \times C}$, where $H$ and $W$ are the height and width of overlapping regions. The error map is defined as $e_{i,j} = ||F_1^{ov}(i,j) - F_2^{ov}(i,j)||^2_2$, where $i, j$ are indices in the error map.}
In each iteration, the choice of the patch is determined by the conditional distribution that measures the similarity of the overlapping region of the synthesized output and the candidate patch.
With a picked patch, the minimum cost path along the overlapping region gives the boundary, and the patch is pasted onto the output texture.
\begin{figure}[htbp]
	\centering
	\includegraphics[width=.8\linewidth]{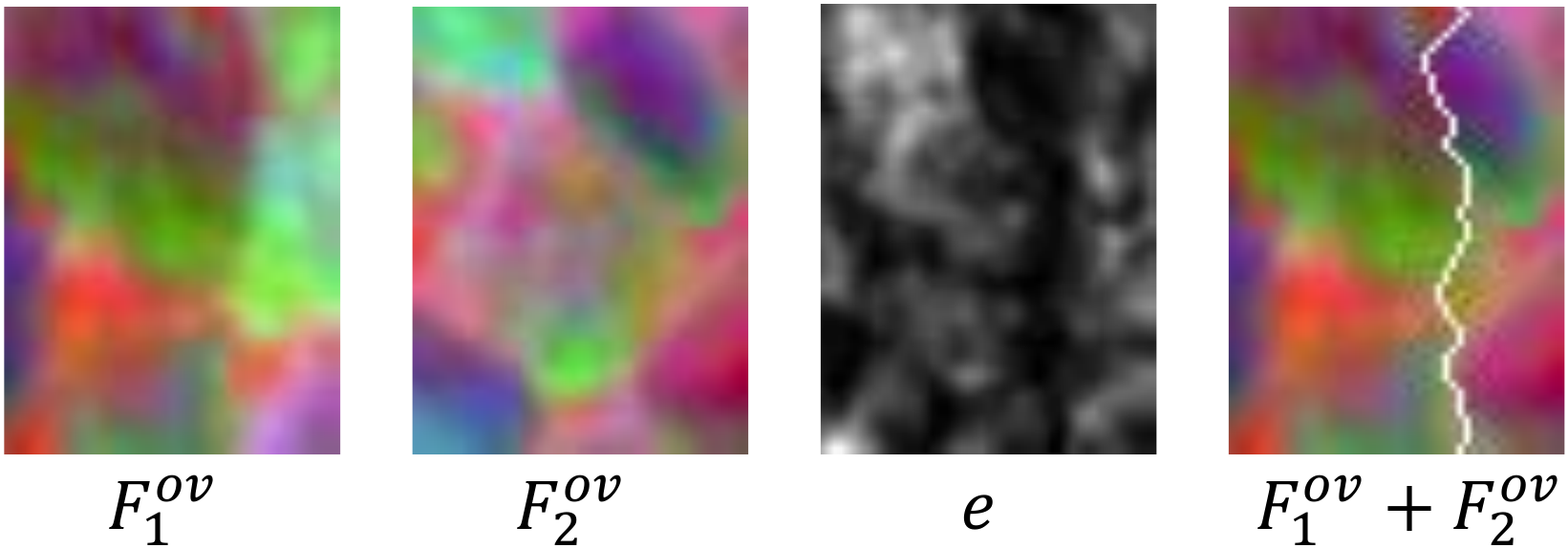}
	\vspace{-2mm}
	\caption{\textbf{Patch Quilting.} The overlapping regions $F^{ov}_1$ and $F^{ov}_2$ are stitched together as $F^{ov}_1 + F^{ov}_2$ based on the minimum cost path along the error map $e$. The cutting path is marked in white.}
\label{fig:quilt}
\end{figure}
\yh{To determine the optimal vertical cut from bottom to top, we leverage dynamic programming to compute the minimum cumulative error $E$ associated with each possible cut position:
\begin{equation}
    E_{i,j} = e_{i,j} + \min\left(E_{i-1,j-1}, E_{i-1,j}, E_{i-1,j+1}\right)
\end{equation}
We then utilize the values of $E$ to backtrack and identify the position of the optimal cut, as illustrated in Fig.~\ref{fig:quilt}.
The process of implicit patch matching for NeRF-based texture synthesis is demonstrated in Alg.~\ref{alg:synthesis}.
\begin{algorithm}
\small
\caption{Implicit Patch Matching\label{alg:synthesis}}
     \begin{flushleft}
        \textbf{Input:} $\{F_i\}$: implicit exemplars \\
        \textbf{Output:} $\hat{F}$: synthesized implicit texture \\
        \textbf{Procedure:}
     \end{flushleft}
    \begin{algorithmic}[1]
        \State Randomly paste a seed patch from $\{ F_i \}$ to the top left of $\hat{F}$
        \Repeat
            \State Determine synthesizing and overlapping regions  $\hat{F}_1,F_1^{ov}$
            \State Pick $K'$ patches $\{ F'_k \}$ with lowest average error $\overline{e}$
            \State Calculate the probability $\{ p'_k \}$ of patches based on $\overline{e}$
            \State Sample a patch from the distribution $\{ p'_k \}$
            \State Compute the cutting edge with minimum cost
            \State Paste the patch to the output texture
        \Until Finish
        \State Return $\hat{F}$
    \end{algorithmic}
\end{algorithm}}

\begin{figure*}[htbp]
	\centering
	\includegraphics[width=1\linewidth]{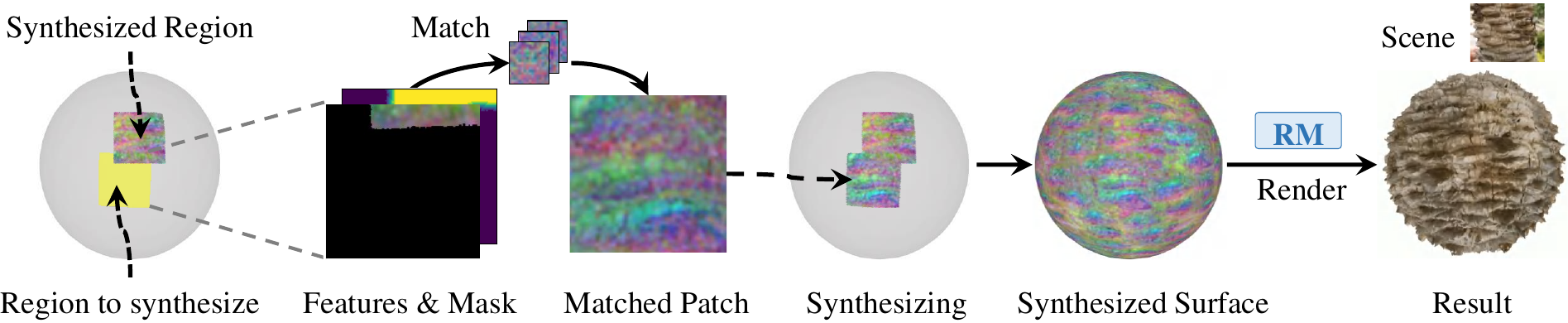}
	\vspace{-6mm}
	\caption{\yh{\textbf{Synthesis on Arbitrary Surfaces.} The patch-matching synthesis algorithm of our NeRF-based texture can also be extended to arbitrary surfaces. Synthesized implicit texture is passed to the Rendering Module (RM) to produce final images of textured surface.}}
\label{fig:synthesis_on_surface}
	\vspace{-2mm}
\end{figure*}

\yh{
\subsubsection{Synthesis on Curved Surface}
By parameterizing local regions on a mesh surface as small rectangles~\cite{turk2001texture,wei2001texture}, \yln{the} synthesis algorithm of our NeRF-based texture representation can also be extended to arbitrary curved surface just like in image texture~\cite{praun2000lapped,soler2002hierarchical}. 

However, our NeRF-based texture synthesis on arbitrary surfaces poses several challenges. Firstly, unlike 2D textures, NeRF-based textures exhibit higher-frequency appearance and geometry, necessitating a higher synthesis resolution. To address this issue, we take measures at both the source and target ends of texture synthesis. Specifically, we perform more (8000) patch samplings from the source scene, each at a higher resolution (128$\times$128) than those used in 2D texture synthesis. This ensures patch matching accuracy and enhances texture representation quality. On the target end, we increase the resolution of the target domain for synthesis.  Previous works~\cite{turk2001texture,wei2001texture} synthesized 2D textures on mesh vertices; however, to preserve the details of NeRF-based textures, these methods would require a mesh of very high resolution to preserve the details of NeRF-based textures, making the real-time rendering of NeRF-based texture intractable. 
\yln{To address this,}
instead of synthesizing texture on mesh vertices, we first parameterize the mesh as an atlas on a high resolution UV map (2048$\times$2048) using existing approaches~\cite{levy2002least,sander2001texture}. Next, we back-map the \yln{grid points} of the UV map to 3D space and synthesize textures on them, obtaining a UV feature map. During the rendering process, query points are projected to the base mesh to obtain their footprints. The UV coordinates of these footprints are calculated through barycentric weighting of underlying triangles and the latent feature is fetched via bilinear interpolation of the UV feature map.

Secondly, texture synthesis on arbitrary surfaces cannot rely on a pre-built \yln{kd}-tree as done in synthesis \yl{in} UV space to accelerate the patch matching process, since the overlapping regions are not fixed. Furthermore, the high resolution and large number of patches further slow down the matching process. To overcome this challenge, we propose to build a multi-resolution pyramid for each patch and \yln{perform} the matching process in a coarse-to-fine manner. Specifically, given a patch region with synthesized parts, we first build a pyramid for it and then identify the most similar patches at the coarsest level while filtering out the remaining patches. The picked patches are then matched at a finer level and re-filtered. At the finest level, the full resolution is used to select the optimal patch for quilting and pasting. This strategy allows us to avoid performing patch matching for the full resolution for the entire patch collection, significantly reducing the time consumption of 
\yln{each} patch matching from approximately 10 seconds to less than 0.05 seconds.

\yln{To cope with anisotropic textures}, given a target surface represented by a triangular mesh, we determine the vector field on it by interpolating \yln{some} user-defined vectors on several control vertices, as proposed in previous works~\cite{turk2001texture,wei2001texture}. Notice that the vector field can also be derived from other state-of-the-art methods~\cite{zhang2006vector,sharp2019vector}. As shown in Fig.~\ref{fig:synthesis_on_surface}, our synthesis algorithm follows the spirit of patch matching and fills the texture space by iterating the following actions until all vertices are grown with textures:

\begin{enumerate}
    \item \textbf{Region Picking.} Randomly pick an unsynthesized vertex with an appropriate distance from the synthesized region to grow texture.

    \item \textbf{Feature Fetching.} With the picked vertex and its orientation vector, place a patch template on it and fetch synthesized features and mask. The fetching process is illustrated in Fig.~\ref{fig:feature_fetch}.

    \item \textbf{Feature Matching and Quilting.} Select a patch that has minimum overlapping error with synthesized regions from candidate patches. Quilt the selected patch and synthesized regions with Alpha blending~\cite{praun2000lapped} or minimum cut quilting~\cite{efros2001image}.

    \item \textbf{Patch Pasting.} With the textured patch template, features of \yln{the} UV grid are calculated via barycentric interpolation. Features on the UV \yln{grid} are then filled in the UV feature map. The pasting process is demonstrated in Fig.~\ref{fig:patch_paste}.
\end{enumerate}

\begin{figure}[htbp]
	\centering
	\includegraphics[width=1.\linewidth]{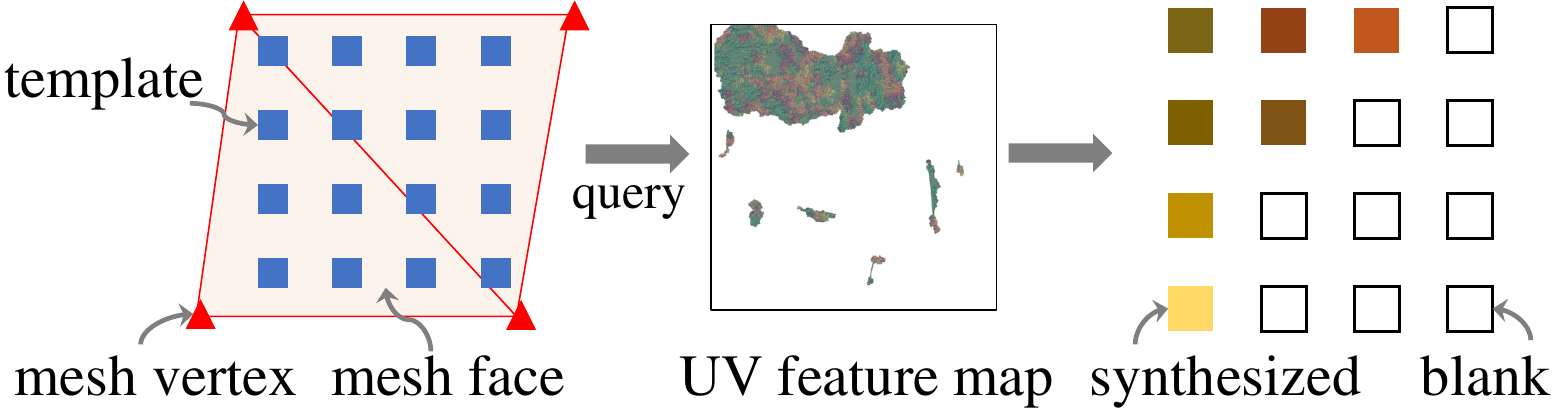}
	\vspace{-4mm}
	\caption{\yh{\textbf{Feature Fetching.} The synthesized features and mask are obtained by \yln{querying} the synthesizing UV feature map with \yln{UV} coordinates of the template patch.}}
\label{fig:feature_fetch}
\end{figure}

\begin{figure}[htbp]
	\centering
	\includegraphics[width=1.\linewidth]{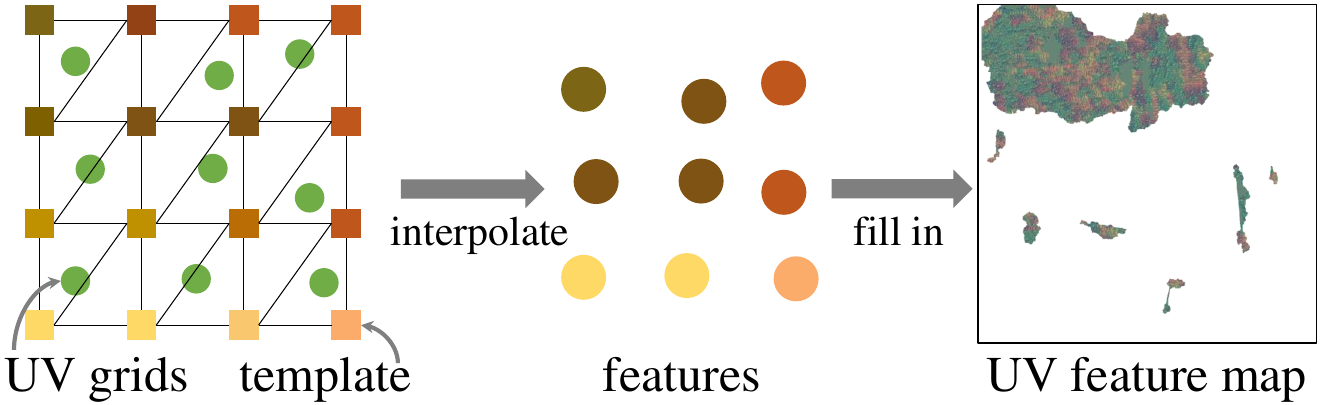}
	\vspace{-4mm}
	\caption{\yh{\textbf{Patch Pasting.} The features of \yln{the} UV grid are calculated via barycentric weighting on the template patch and the synthesized results are filled in the UV feature map.}}
\label{fig:patch_paste}
\end{figure}

}

\begin{figure*}[tbp]
	\centering
	\includegraphics[width=1.0\linewidth]{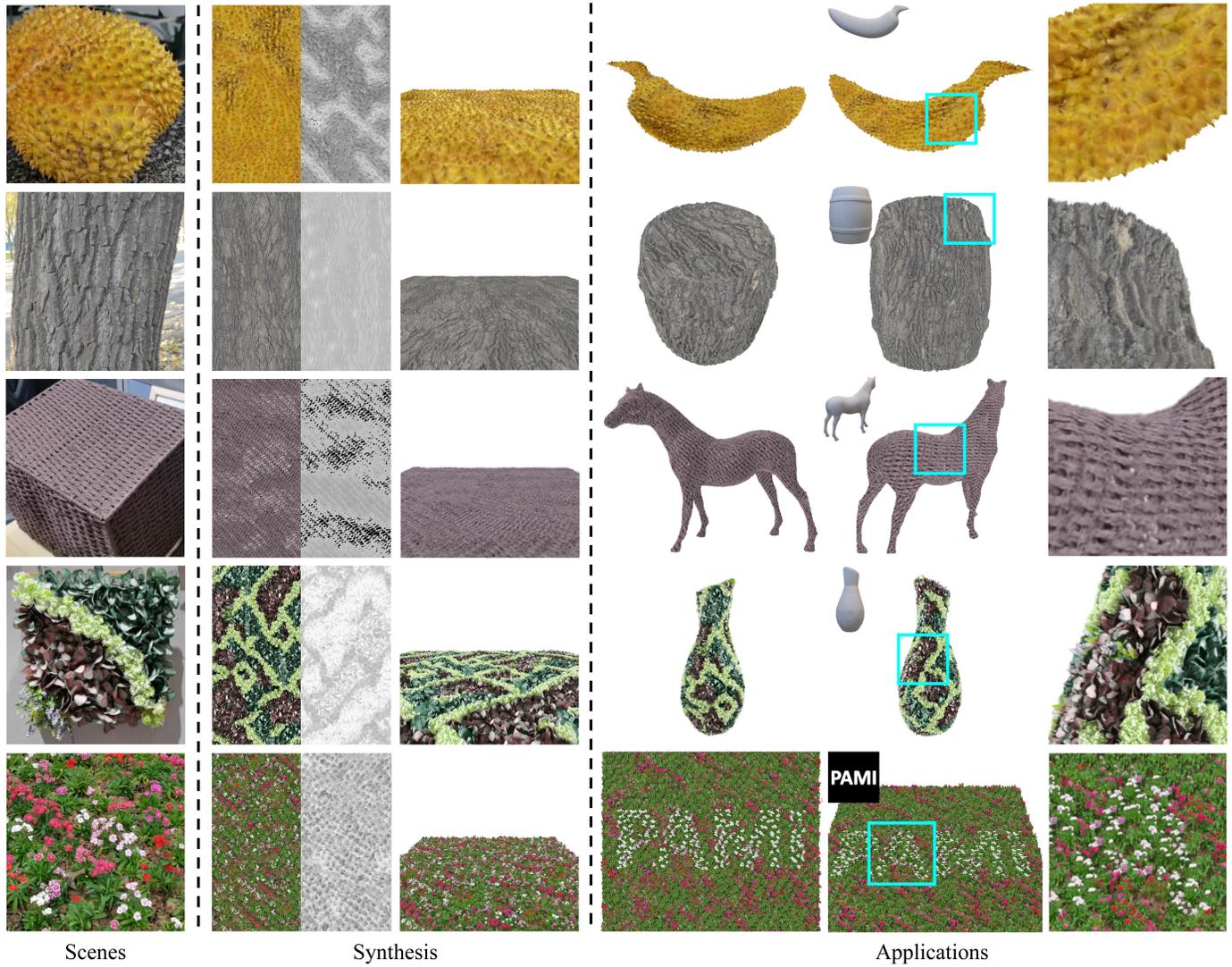}
	\vspace{-7mm}
	\caption{\textbf{Texture Synthesis and Applications.} We show the synthesized textures of durian, tree bark, woven basket, leaves, and flowers. The textures are also applied to different shapes. The last example is constrained synthesis guided by the text image.}
\label{fig:application}
\end{figure*}

\begin{figure*}[tbp]
	\centering
	\includegraphics[width=1\linewidth]{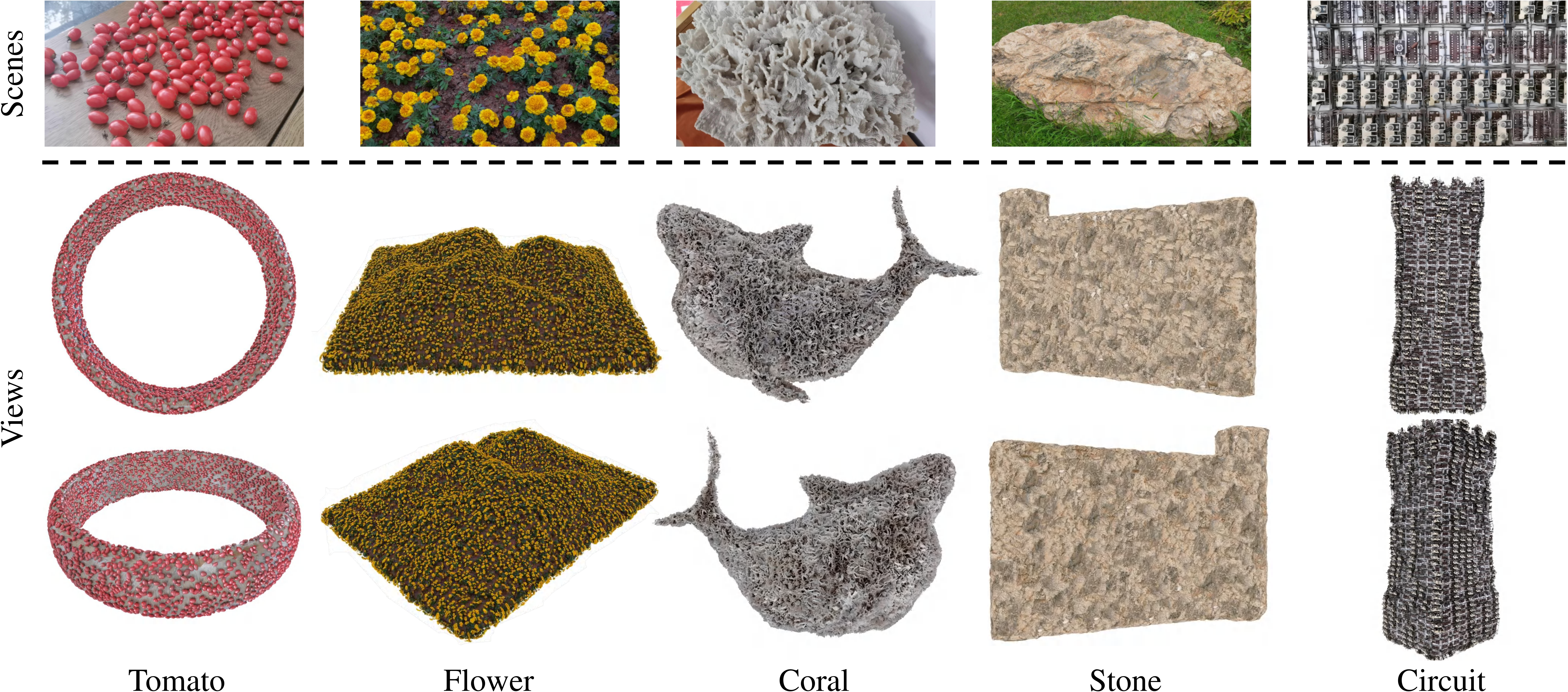}
	\vspace{-6mm}
	\caption{\yh{\textbf{Texture \yln{Synthesis on Curved Surfaces}.} Our synthesis algorithm can be extended to arbitrary curved surfaces, considering the continuity on the surface instead of UV space.}}
\label{fig:curved_results}
\end{figure*}

\begin{figure*}[tbp]
	\centering
	\includegraphics[width=1\linewidth]{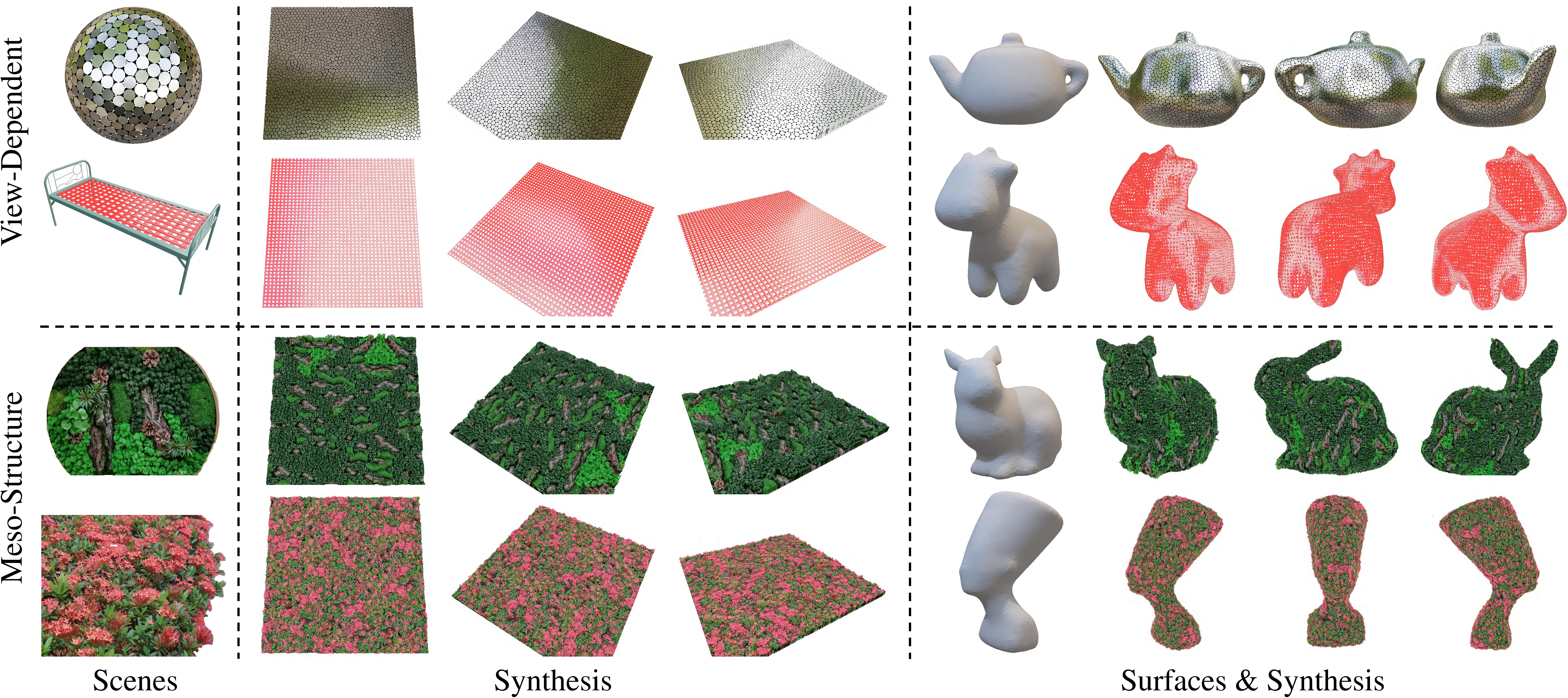}
	\vspace{-6mm}
	\caption{\hyh{\textbf{Synthesis of View-Dependent and Meso-Structure Textures.} Our NeRF-based approach synthesizes textures that capture both the view-dependent appearance (in the first two rows) and meso-structure details (in the last two rows) with accuracy and fidelity. Our method preserves these attributes of textures and achieves high-quality rendering results.}}
\label{fig:more_exp}
\end{figure*}

\subsubsection{Latent Feature Clustering}
Ideally, the metric of latent space should be consistent with that of the reconstructed content space to ensure the plausibility of patch matching.
Thanks to the continuity of neural networks, latent features close to each other reconstruct similar textures.
However, it does not guarantee that similar latent features represent similar texture contents.
To this end, we ensure the consistency of metrics in two aspects.
First, latent features corresponding to similar texture contents have similar optimization targets (e.g. $k_d$, $k_s$, $g$ and $\theta$) during the training, which means that they have close optima when the training converges.
Second, to avoid the latent features corresponding to similar textures falling into different optima during training, we introduce a clustering loss~\cite{xie2016unsupervised} for latent features into the optimization objective.
Student's $\tau$-distribution is used as the kernel to measure the similarity~\cite{van2008visualizing} between latent features $f_i$ and trainable cluster centers $\mu_j$.
The distribution $Q$ and its hardened auxiliary distribution $P$ are defined as:
\begin{equation}
    q_{ij} = \frac{(1+||f_i-\mu_j||^2_2 / \kappa)^{-\frac{\kappa+1}{2}}}{\sum\limits_{j'}(1+||f_i-\mu_{j'}||^2_2 / \kappa)^{-\frac{\kappa+1}{2}}}, \ 
    p_{ij}=\frac{q_{ij}^2 / \sum\limits_i q_{ij}}{\sum\limits_{j'}(q_{ij'}^2 / \sum\limits_i q_{ij'})}
\end{equation}%
where $\kappa$ is the degree of freedom of the Student's $\tau$-distribution.
$P$ is stricter than $Q$ and closer to 0 or 1.
The clustering loss is given by the KL divergence~\cite{kullback1951information} between them: $L_{clu}=KL(P||Q)$.
For hash grids at each resolution level, we cluster the embedding features with the clustering loss.

\subsubsection{Texture Mapping}
Given a new 3D shape with known UV coordinates, query point $x$ is projected onto the surface, with the foot point denoted as $\tilde{x}_c$, as described in Sec.~\ref{subsec:itr}.
The latent features $\tilde{f}(x)$ of the $x$ is obtained by bilinear interpolation on the synthesized texture with UV coordinates of $\tilde{x}_c$.
The residual transformation from the original coarse mesh local frame to the sampling local frame  $T_s^{-1}(x)T_c(x)$ is also obtained by nearest-neighbor interpolation on synthesized $T_s$ and $T_c$ maps.
Based on the feature and SDF value, the network predicts the appearance and geometry of the query point.
With the transformation of the new tangent space on the target surface, denoted as $\tilde{T}_c(x)$, the predicted normal on the new shape is calculated as:
\begin{equation}
\begin{aligned}
    \tilde{n}_f(x)=\tilde{T}_c(x)T_s^{-1}(x)T_c(x)R(\theta(x), \phi(x)), \\ 
    R(\theta, \phi) = (\sin\theta \cos\phi, \sin\theta \sin\phi, \cos\theta)^T
\end{aligned}
\end{equation}
The density and reflection coefficients are also calculated by $\tilde{f}(x)$ and SDF value $\tilde{s}(x)$ relative to the new shape.

\begin{table*}[ht]
  \centering
  \caption{\textbf{Quantitative comparison of view synthesis.} We show the average PSNR/SSIM/LPIPS for novel view synthesis on DTU.}\label{tab:recon}
  \vspace{-2mm}
  \setlength\tabcolsep{0pt}
    \begin{tabular*}{\linewidth}{@{\extracolsep{\fill}} lSSSSSSSSSSSS }
        \toprule[1pt]
        \multirow{2}{*}{Methods} &
          \multicolumn{3}{c}{\textbf{Scan 55}} &
          \multicolumn{3}{c}{\textbf{Scan 83}} &
          \multicolumn{3}{c}{\textbf{Scan 105}} &
          \multicolumn{3}{c}{\textbf{Scan 122}} \\
          \cmidrule(lr){2-4}
          \cmidrule(lr){5-7}
          \cmidrule(lr){8-10}
          \cmidrule(lr){11-13}
          & {\textbf{PSNR($\uparrow$)}} & {\textbf{SSIM($\uparrow$)}} & {\textbf{LPIPS($\downarrow$)}} & {\textbf{PSNR($\uparrow$)}} & {\textbf{SSIM($\uparrow$)}} & {\textbf{LPIPS($\downarrow$)}} & {\textbf{PSNR($\uparrow$)}} & {\textbf{SSIM($\uparrow$)}} & {\textbf{LPIPS($\downarrow$)}} & {\textbf{PSNR($\uparrow$)}} & {\textbf{SSIM($\uparrow$)}} & {\textbf{LPIPS($\downarrow$)}} \\
          \midrule
        NeRF     
        & {28.244} & {0.940} & {0.212}      
        & {37.816} & {0.990} & {0.092}
        & {34.152} & {0.947} & {0.208}
        & {36.464} & {0.979} & {0.135}\\
        NGP     
        & \textbf{34.108} & \textbf{0.991} & \textbf{0.086} 
        & \underline{42.602} & \underline{0.996} & \underline{0.049} 
        & \textbf{38.247} & \textbf{0.991} & \underline{0.085} 
        & \underline{41.976} & \underline{0.996} & \underline{0.057} \\
        Ours                               
        & \underline{32.378} & \underline{0.988} & \underline{0.104}
        & \textbf{43.842} & \textbf{0.998} & \textbf{0.027}
        & \underline{36.809} & \underline{0.990} & \textbf{0.067} 
        & \textbf{42.704} & \textbf{0.998} & \textbf{0.031} \\
        \bottomrule[1pt]
      \end{tabular*}
\end{table*}

\subsection{Optimization}
\label{subsec:optimization}
Our model is trained with the Adam optimizer~\cite{kingma2014adam}.
The optimization target of our method consists of four terms: 
$L = L_{rec} + \lambda_1 L_{clu} + \lambda_2 L_{dis} + \lambda_3 L_{nor}$.
$L_{rec}$ is the L1 RGB reconstruction loss. $L_{dis}$ is the distortion loss~\cite{barron2022mip} removing floating artifacts.
$L_{nor}$ supervises the prediction of $(\theta, \phi)$ based on the negative gradients of density $\sigma(x)$ relative to $x$. Owing to the noise of density gradients, we employ the relaxed cosine distance to supervise the estimated normal:
\begin{equation}
    L_{nor} = - \cos\{\min(\langle<-\frac{\mathrm{d}\sigma(x)}{\mathrm{d}x}, n_f(x)\rangle>, \frac{\pi}{8})\}
\end{equation}
In our experiments, $\lambda_1$, $\lambda_2$, and $\lambda_3$ are set to $10^{-5}$, $10^{-2}$, and $1$.


\section{Results}
\label{sec:results}

In this section, we perform several experiments to demonstrate the utility of our method. We will firstly show the results on texture synthesis and applications in Sec.~\ref{subsec:synthesis:app}. Then we quantitatively and qualitatively compare the novel view synthesis quality to show the rendering quality of our method in Sec.~\ref{subsec:view}. We also compare the 2D texture and our representation in Sec.~\ref{subsec:vs2d}, which  the advantage of our method in texture modeling.
Finally, we compare with NeRF-Tex in Sec.~\ref{subsec:vs_nerftex} and perform an ablation study on the impact of latent feature constraint in Sec.~\ref{sec:ablation}.

\subsection{Texture Synthesis and Applications}
\label{subsec:synthesis:app}
We demonstrate the utility of our method by acquiring and synthesizing textures from the real world captured by a mobile phone as shown in Fig.~\ref{fig:application}.
The target texture includes durian,  bark, fabric, leaves, and flowers.
The synthesized results and depth visualization are shown in the $2${nd} and $3${rd} columns.
We also applied captured textures to grow on the desired shape or pattern shown in  $4$th-$6${th} columns.
We synthesize the durian's texture with thorns and transfer it to a banana.
The tree bark is usually covered with stripes of ravines. We synthesize and apply such texture to a barrel shape and obtain a wooden barrel.
We capture fabric texture on a woven basket and construct a woven horse.
Leaves and grass are also typical textures in nature; we synthesize  ocean of leaves and grass and apply it to a vase.
We also synthesize colorful flowers guided by the shown text image, by considering the rendered color of patches during texture synthesis (see supplementary for details).
The zoomed-in view in  $6${th} column shows the effect of the material on oblique views and object edges, where 2D textures appear unrealistic due to the lack of meso-structure modeling, demonstrating the advantages of our method over 2D textures.

\yh{We also show the results of our NeRF-based texture synthesis on curved \yln{surfaces} using various captured textures in Fig.~\ref{fig:curved_results}. We capture scenes of tomatoes on the desk, flowers on the ground, a coral, a stone, and circuits of an old computer. These textures are synthesized on surfaces of a ring, a mountain-valley shape, a shark, a wall, and a tower respectively. The synthesis only considers the \yln{neighboring} relationship in the 3D space rather than the UV space.}

\hyh{
To better highlight the advantages of our NeRF-based method in modeling view-dependent appearance and meso-structure geometry, we present additional results in Fig.~\ref{fig:more_exp}. The first two rows showcase the texture synthesis of mirror balls and a metal bed, which are made up of reflective mirror faces and metal specular materials, respectively. To improve the modeling of such specular materials and accurately reflect high-frequency environment lights, we use the lighting representation of Ref-NeRF~\cite{verbin2022ref} by introducing an extra MLP to predict the reflected color of a reflective direction during the shading step. We also synthesize textures on Utah Teapot~\cite{torrence2006martin} and Spot~\cite{crane2013robust}, resulting in high-fidelity view-dependent appearance of the textured meshes. 
In addition, we synthesize textures of various leaves, barks, pinecones, moss, and tiny flowers, as shown in the last two rows. We also evaluate our method on the Standord Bunny~\cite{turk1994zippered} and Nefertiti, demonstrating that the NeRF-based representation preserves the high fidelity of captured textures in both view-dependent appearance and meso-scale geometry.

}

\begin{figure}[htbp]
	\centering
	\includegraphics[width=1\linewidth]{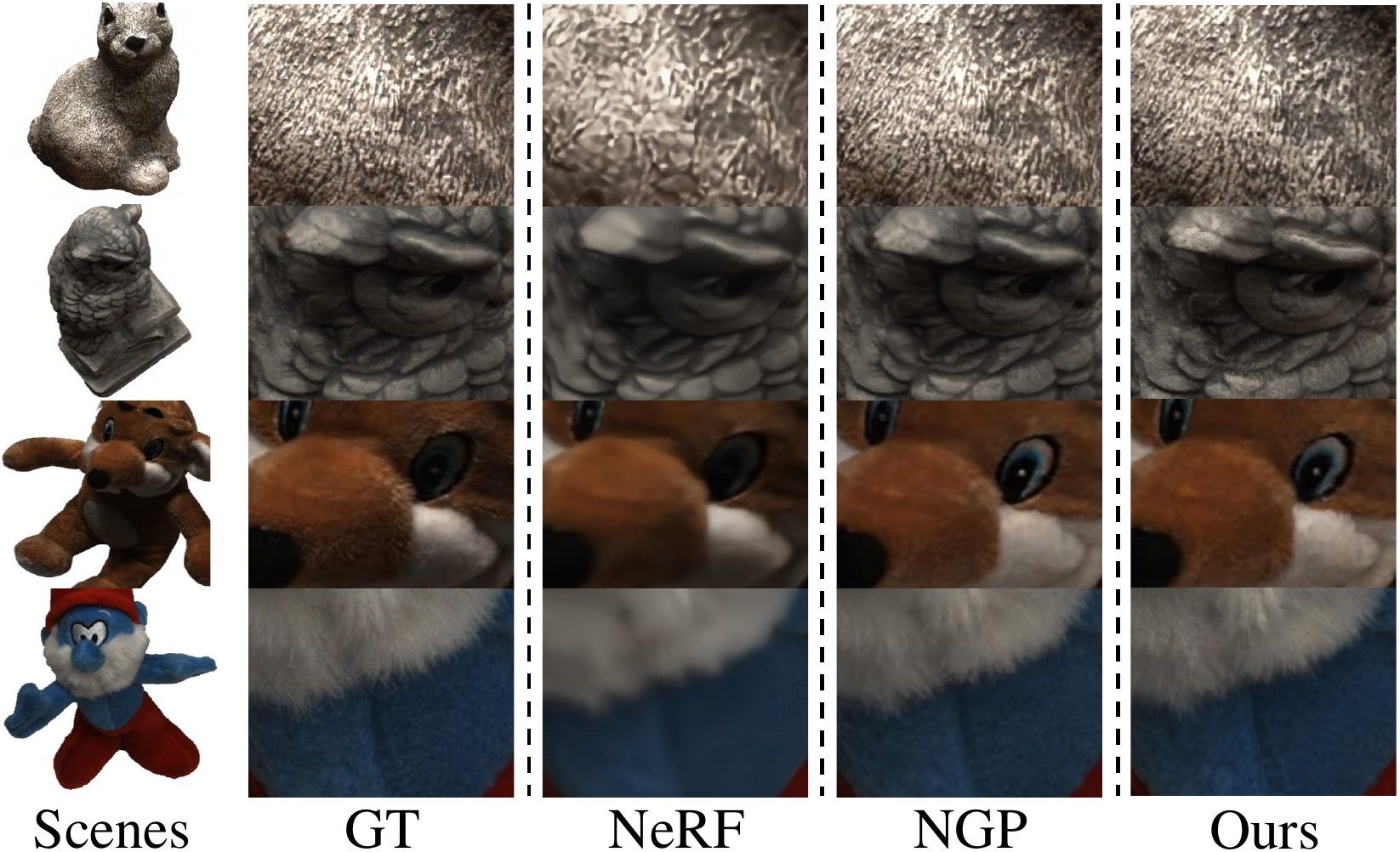}
	\vspace{-2mm}
	\caption{\textbf{Qualitative Comparison of View Synthesis \yln{Results}.} Note that our method supports texture capture, synthesis and application while visually close to the state of the arts.}
\label{fig:recon}
\end{figure}

\subsection{View synthesis quality}
\label{subsec:view}
We evaluate the view synthesis quality of our method on the published dataset
DTU~\cite{aanaes2016large}, in which the scenes are of objects suitable for our method to represent as they contain texture-like structure. 
We test on 4 scenes with masks provided by \cite{yariv2020multiview}.
In each scene, 5 images are randomly picked as the test set.
Qualitative comparison with NeRF~\cite{mildenhall2021nerf}, Instant-NGP (NGP)~\cite{mueller2022instant} and ours is shown in Fig.~\ref{fig:recon}.
We report the PSNR, SSIM, and LPIPS in Tab.~\ref{tab:recon}.
Due to the specific design for disentangling meso-structure and materials, our approach is slightly worse than NGP in some quantitative comparisons.
Despite this, our rendered results are still realistic in high-frequency details and perceptually close to NGP's results.

\subsection{Comparison with 2D texture}
\label{subsec:vs2d}
To verify the advantages of our texture representation over 2D image textures, we conduct quantitative and qualitative experiments to demonstrate it.
To obtain 2D texture patches, we simultaneously render the RGB patches when sampling patches as described in Sec.~\ref{subsec:its}.
Based on the RGB patches, we use the patch matching algorithm to synthesize a texture image the same size as our generated neural texture.
We render both 2D and neural textures in different angles of elevation from $0^\circ$ to $80^\circ$ as samples for comparison.

\begin{figure}[htbp]
	\centering
	\includegraphics[width=1\linewidth]{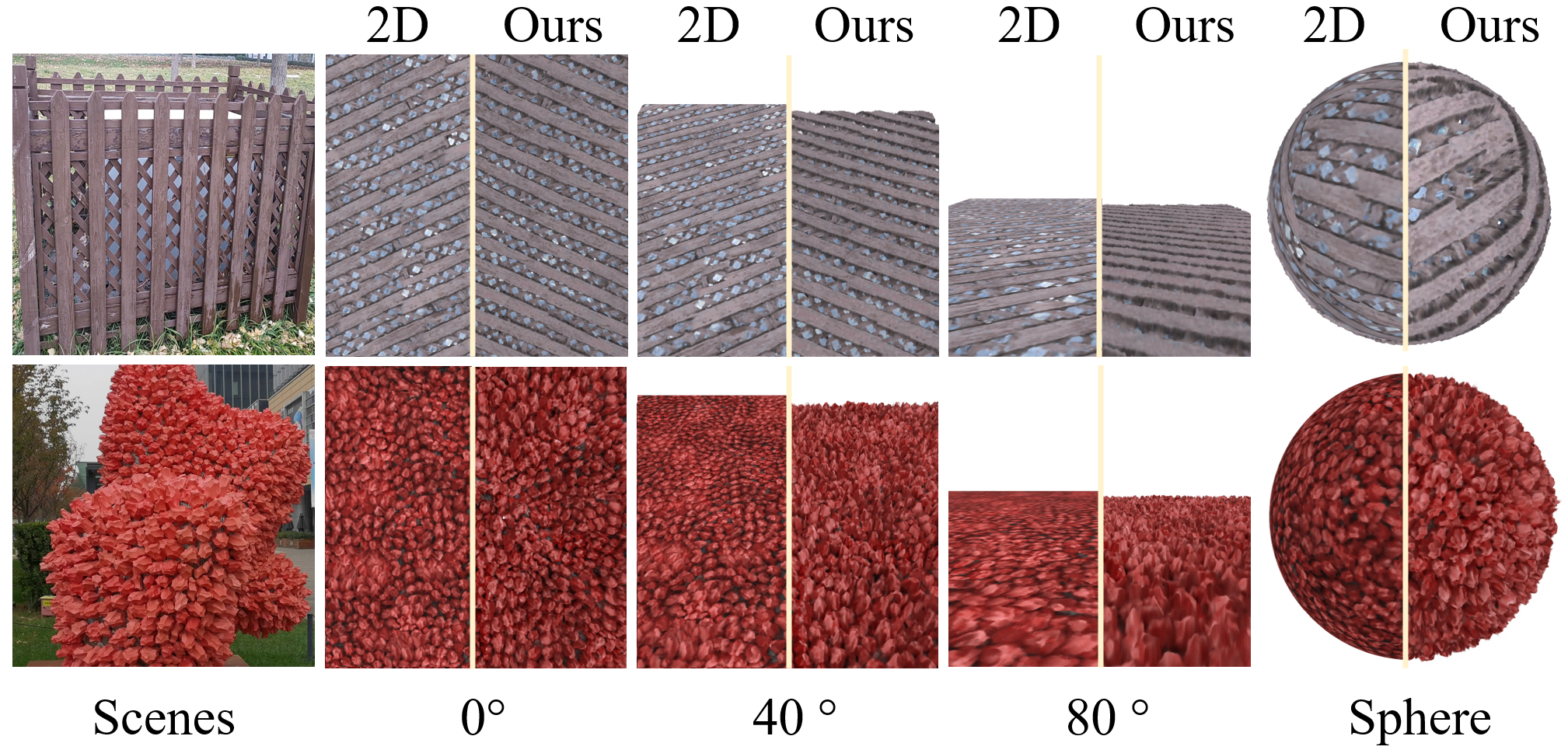}
	\vspace{-7mm}
	\caption{\textbf{Qualitative Comparison with 2D Textures.} We show the rendering results of our synthesized textures and 2D textures. Our representation of maintains realism even at high-elevation viewing angles.}
\label{fig:vs2d}
\end{figure}

\noindent\textbf{Single Image Fr{\'e}chet Inception Distance (SIFID)} 
SIFID introduced in \cite{shaham2019singan} is a commonly used metric to assess the realism of generated images. 
We crop the regions, where the corresponding 3D shape approximate planes, from the captured images as ground truths.
We then calculate the SIFIDs between rendered textures with ground truths of the closest viewing directions relative to planes.
Average SIFIDs reported in Tab.~\ref{tab:sifids} indicate that our texture representation is more realistic than 2D textures.

\begin{table}[htbp]
  \caption{\textbf{Quantitative Comparison with 2D Texture.} Our texture has lower SIFIDs in all elevation angles.}\label{tab:sifids}
	\vspace{-2mm}
  \centering
  \setlength\tabcolsep{0pt}
    \begin{tabular*}{\columnwidth}{@{\extracolsep{\fill}} lSSSSSS }
        \toprule[1pt]
          Degree & \textbf{$0^\circ$}& \textbf{$20^\circ$}& \textbf{$40^\circ$} & \textbf{$60^\circ$} & \textbf{$80^\circ$} & \textbf{Average} \\
          \midrule
        2D     & {$0.73$} & {$0.75$} & {$0.82$} & {$1.21$} & {$1.75$} & {$1.05$} \\
        Ours   & \textbf{0.52} & \textbf{0.51} & \textbf{0.56} & \textbf{0.58} & \textbf{0.82} & \textbf{0.60} \\
        \bottomrule[1pt]
         \end{tabular*}
\end{table}

\noindent\textbf{Qualitative comparison}
We also show the qualitative comparison of 2D image texture with our representation in Fig.~\ref{fig:vs2d} in different viewing directions. The synthesized 2D texture of meso-structure will be unrealistic at high elevation angles while our representation can well represent the geometry occlusion of meso-structure.

\subsection{Comparison with NeRF-Tex}
\label{subsec:vs_nerftex}
\begin{figure}[htbp]
	\centering
	\includegraphics[width=1\linewidth]{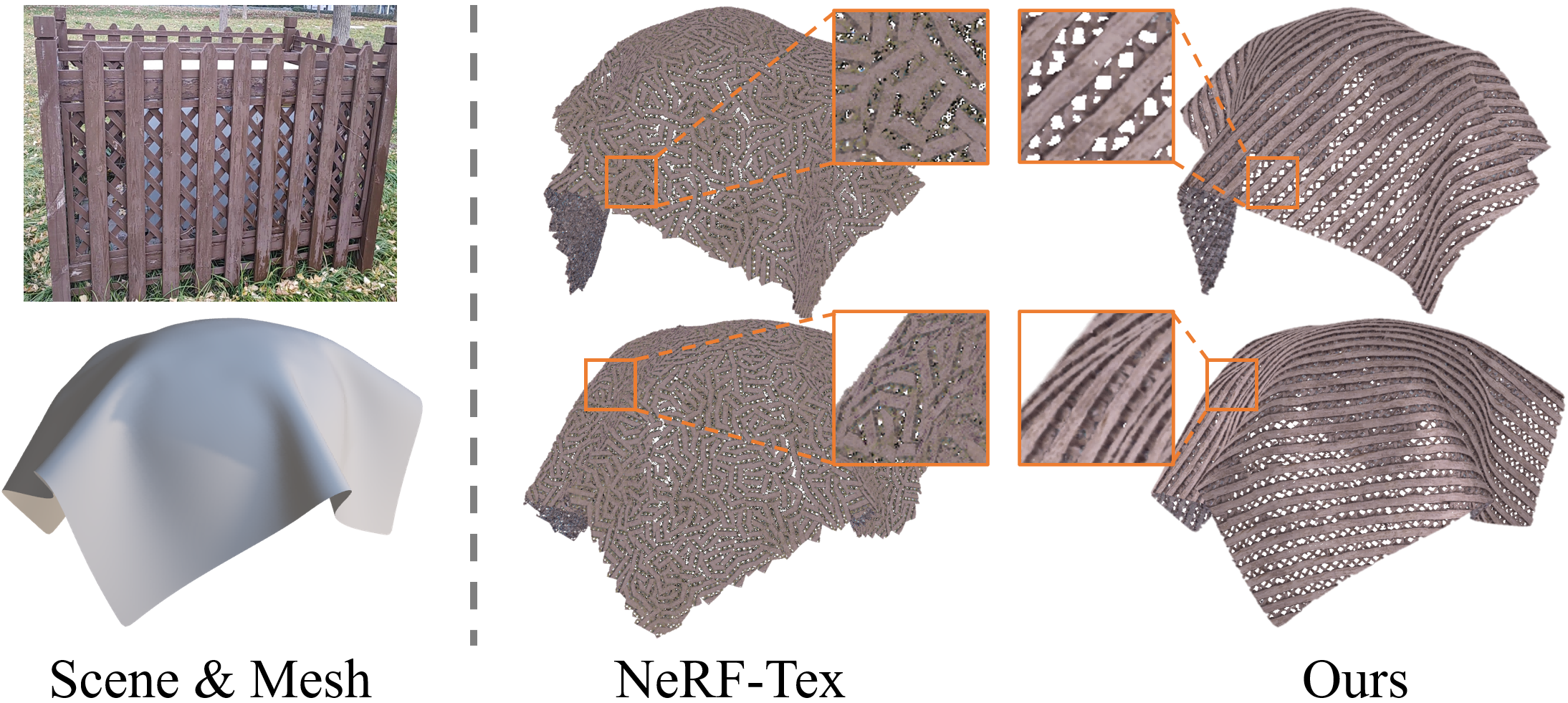}
	\vspace{-6mm}
	\caption{\textbf{Comparison with NeRF-Tex~\cite{baatz2021nerf}} Our method demonstrates superior preservation of texture continuity and structure thanks to the synthesis algorithm of NeRF texture.}
\label{fig:vs_nerf_tex}
\end{figure}

We present a visual comparison between NeRF-Tex~\cite{baatz2021nerf} and our proposed method, as demonstrated in Fig.~\ref{fig:vs_nerf_tex}. In contrast to our approach, NeRF-Tex does not perform texture synthesis; instead, it repeatedly places planar texture patches on anchor points of the mesh in an unstructured manner, leading to a loss of regularity for typical structural textures. Besides, it is crucial to note that NeRF-Tex trains a NeRF using synthetic data with known tightly bound planar geometry, which cannot be directly applied to real-world data. Thus, we utilize our coarse-fine disentanglement representation to generate multi-view images of real-world meso-structure textures within a bounding box, serving as training data for NeRF-Tex.

\subsection{Ablation on Clustering Constraint}
\label{sec:ablation}

\begin{figure}[htbp]
	\centering
	\includegraphics[width=0.95\linewidth]{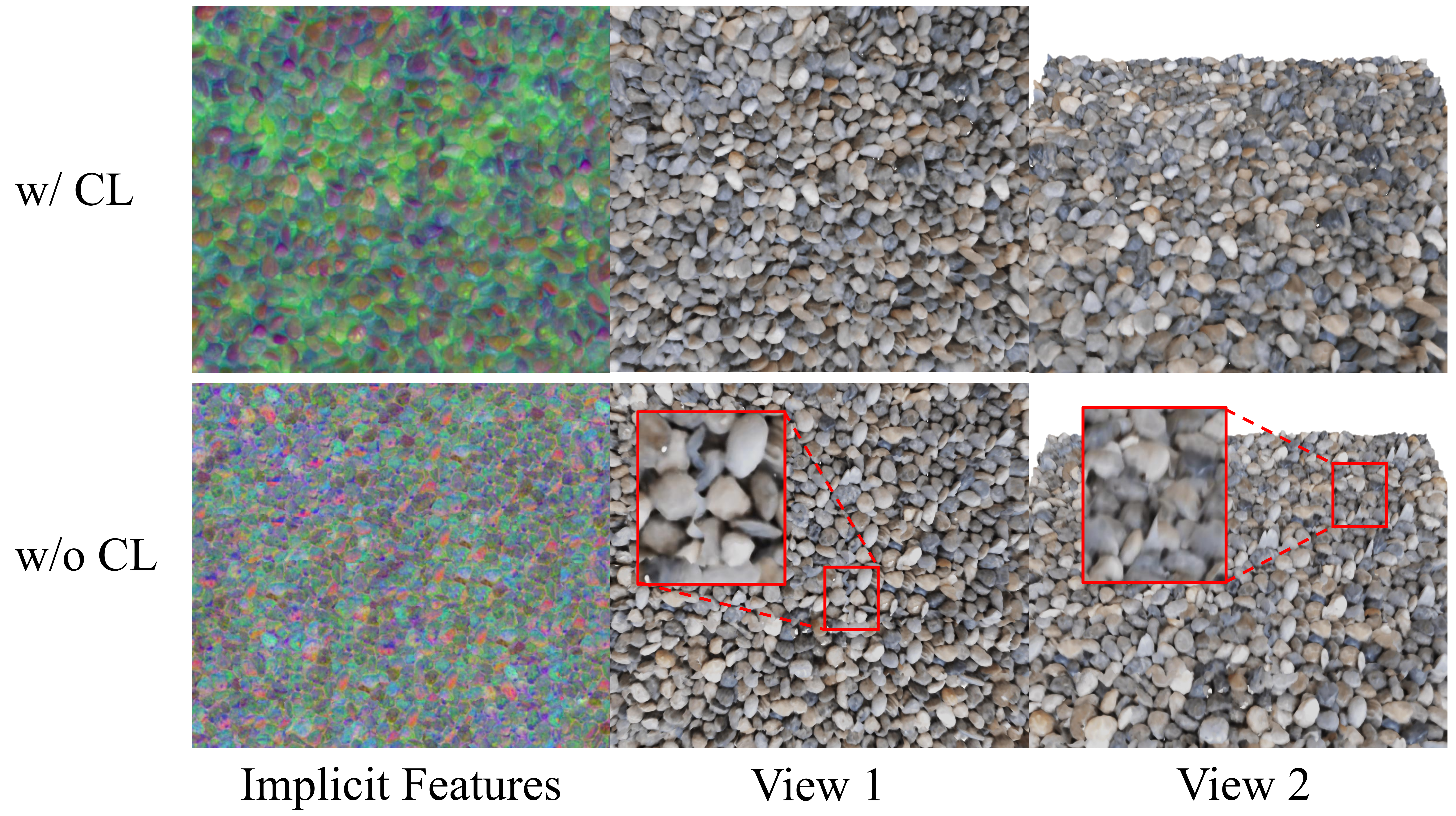}
	\vspace{-2mm}
	\caption{\textbf{Impact of Clustering Constraint.} With the clustering loss (w/ CL), latent features are constrained to cluster, which reduces the distance of latent features corresponding to similar textures and further reduces artifacts in synthesized results.}
\label{fig:clustering}
\end{figure}

The complexity and randomness of textures can easily lead to the disordered distribution of features, even if these features share the same reconstruction target.
The clustering constraint regularizes the latent distribution by encouraging similar textures  represented by close features.
We visualize the synthesized feature  with and without the constraint by Principal Component Analysis dimensionality reduction to 3D, which is further visualized as RGB channels in Fig.~\ref{fig:clustering} (left).
We found that the constraint makes the latent distribution more compact and reduces the variance.
Results without the constraint tend to have more artifacts (Fig.~\ref{fig:clustering} (right)).

\yh{
\subsection{Ablation on Patch Resolution} 
The resolution of patches is a pre-defined hyperparameter; therefore, we perform ablation experiments to analyze its impact.
According to the work~\cite{efros2001image}, patch size needs to be large enough to capture the structure of the texture, yet small enough that the algorithm can stitch them freely. 
With a determined patch size, we found that a patch with a 128$\times$128 resolution consistently yields satisfactory texture reconstructions. Compared with 128$\times$128, \yln{using} the resolution of 256$\times$256 \yln{does not} significantly \yln{improve results} but increases the storage cost. Therefore, in our experiments, we use a resolution of 128$\times$128 to acquire texture patches.}

\begin{figure}[htbp]
	\centering
	\includegraphics[width=1.\linewidth]{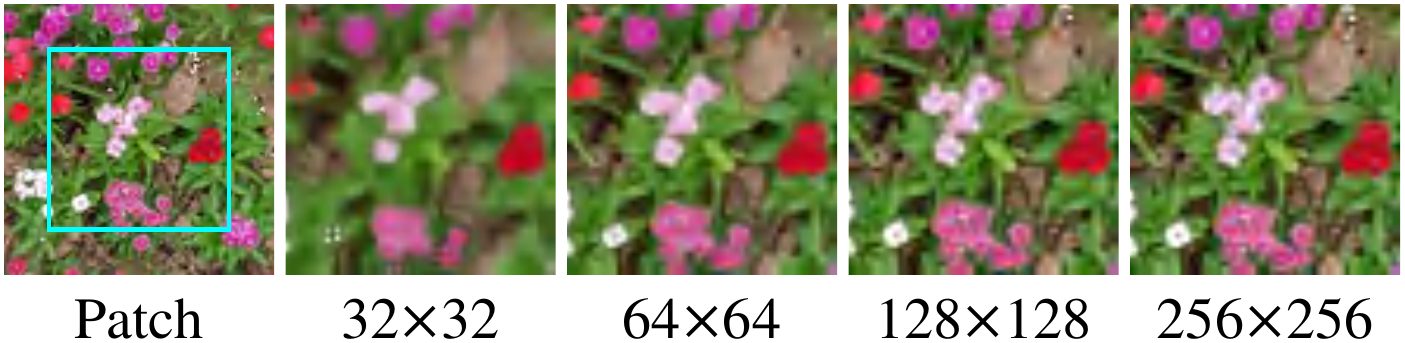}
	\vspace{-2mm}
	\caption{\textbf{Ablation on Patch Resolution.} A resolution of 128$\times$128 is sufficient to represent a patch area of the required size.}
\label{fig:patch_resolution}
\end{figure}

\hyh{

\subsection{Ablation on the Number of Training Views}

NeRF is built on the principle that when light rays intersect a surface at the same position and from the same view direction, 
they have the same color. During training, NeRF casts rays from training views and optimizes rendering loss for each pixel. Therefore, increasing the number of training images can reduce the ambiguity of ray intersection and improve the synthesis of new perspectives. Our method focuses on textured scenes with high-frequency features, where ambiguity is less pronounced compared to scenes that lack color and geometry richness.

We conducted an ablation study on a scene with a ground covered in rocks. The rendering results of our model, trained on different numbers of views, are shown in Fig.~\ref{fig:ablation_views}. With only a few views (e.g., 4 views shown in the first column), the rendering results exhibit blue tint and artifacts, as the training views are unable to capture every detail of the texture. With 16 training views (second column), most parts of the scene become much clearer, but a few floating pieces remain. When using 64 training views, the visual quality is very close to that of using 256 views. This experiment demonstrates that texture reconstruction with a small number of training views remains a challenge for our method. In our experiments, we typically capture the scene with 150 to 300 training views from a short video to ensure the quality of the rendered textures.

\begin{figure}[H]
	\centering
	\includegraphics[width=1.\linewidth]{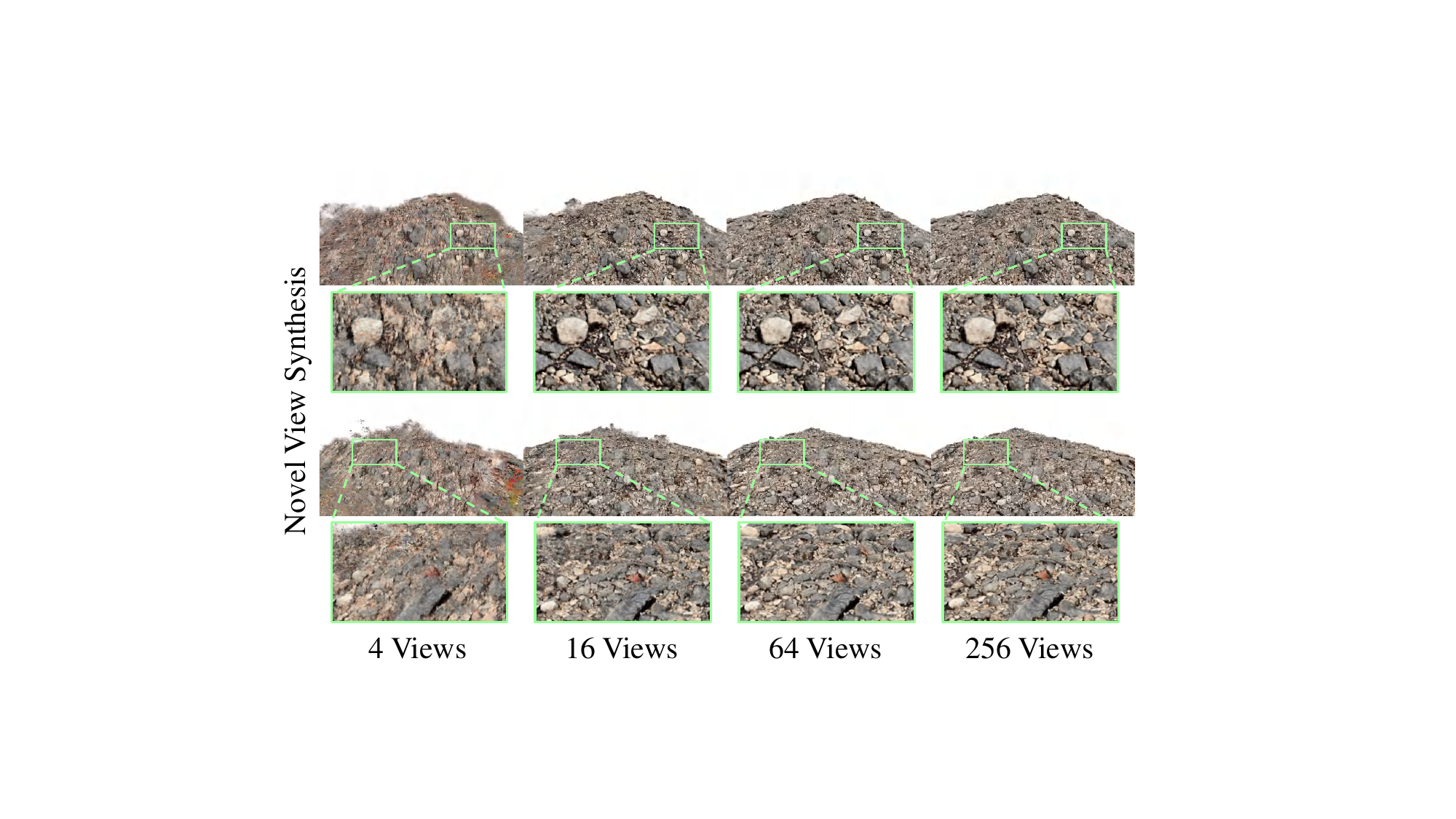}
	\vspace{-2mm}
	\caption{\textbf{Ablation Study on the Number of Training Views.} Using fewer training views results in ambiguity in geometry and appearance. Increasing the number of training views improves the quality of rendered textures.}
\label{fig:ablation_views}
\end{figure}

\subsection{Rendering Speed Analysis}

Our approach utilizes hash grids to efficiently retrieve latent features and employs tiny MLPs~\cite{tiny-cuda-nn} for quick color and opacity querying. However, our method consumes additional time for the K-Nearest Neighbor (KNN) search during coarse normal calculation and ray tracing for footprint projection. To address this, we applied specific engineering strategies to enhance the speed of these two processes in our implementation:
\textbf{1)} To expedite the KNN search, we evenly divide the space into bins~\cite{hoetzlein2014fast} and insert each vertex of the base mesh into them. During the KNN search, we only need to search mesh vertices within neighboring bins of the query point within a specified radius.
\textbf{2)} Additionally, we construct a Bounding Volume Hierarchy (BVH)~\cite{ha2004bvh} with nodes formatted in Axis-Aligned Bounding Boxes (AABBs)~\cite{beckmann1990r} to organize the triangles of the base mesh. The pre-built BVH avoids the unnecessary ray collision detection of those triangles within AABBs that do not intersect, thereby accelerating the ray-tracing process~\cite{meister2021survey}.
All these operations are implemented using CUDA and executed by hundreds of threads in parallel on the GPU, enabling fast calculations.

We evaluated the rendering speed of our method and compared it to Vanilla-NeRF~\cite{mildenhall2021nerf} and Instant-NGP~\cite{mueller2022instant} as benchmarks. The rendering speed, measured in frames per second (FPS), is presented in Tab.~\ref{tab:render_speed}. Although our method is slower than Instant-NGP due to the additional calculations, it still achieves real-time rendering speed and is much faster than Vanilla-NeRF.

\begin{table}[h]
\centering
\caption{\textbf{Rendering speed comparison.} We compared the rendering speed of NeRF, Instant-NGP and our method, at a resolution of $512\times512$.}
 \label{tab:render_speed}
\resizebox{.9\columnwidth}{!}{
\begin{tabular}{ lccc }
\toprule[1pt]
 Method & NeRF~\cite{mildenhall2021nerf} & Instant-NGP~\cite{mueller2022instant} &  Ours \\ \hline 
 Speed (FPS) $\uparrow$ & 0.02 & 307.05 & 84.45 \\ 
 \hline
\end{tabular}}
\end{table}

}


\hyh{

\begin{figure*}[thbp]
	\centering
	\includegraphics[width=0.9\linewidth]{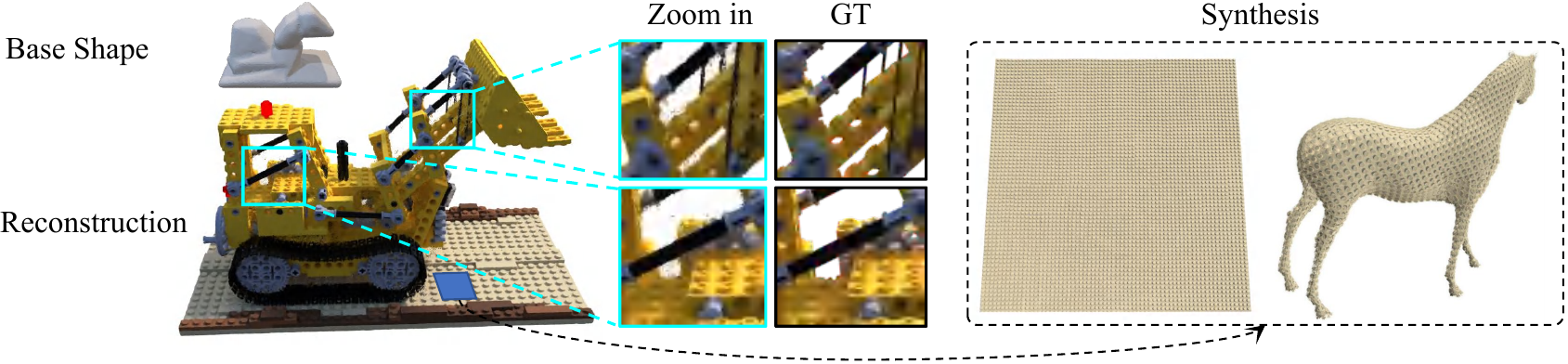}
	\vspace{-2mm}
	\caption{\textbf{Challenges on Texture Capture.} Our approach fails to reconstruct and capture textures on shapes with complex coarse geometry due to the difficulty in base shape estimation and patch sampling on regions with limited spatial extents of the base surface. On the contrary, the bump texture on the Lego base can be easily acquired and synthesized.}
\label{fig:limitation}
\end{figure*}

\section{Challenge Analysis}
\label{sec:limitation}
As we have demonstrated, our method can easily capture, reconstruct, and synthesize common real-world textures on various objects. To better understand and explore the boundaries of the method's capabilities, we conducted several experiments and analyses on more challenging scenes in this section. We summarize the challenges into two aspects: texture capture and texture synthesis, and discuss them respectively in the following.

\subsection{Challenging Texture Capture}
Our approach faces certain challenges when it comes to capturing textures from objects with \textbf{1)} limited coarse extents or \textbf{2)} complex topology.

\noindent\textbf{Limited Coarse Extents} For scenes where the coarse shape has limited spatial extents, the size of sampled patches is also limited. In extreme cases, the patch size will be too small to capture texture patterns, resulting in synthesized textures containing numerous artifacts. As shown in the second row of Fig.~\ref{fig:limitation_capture}, the truss's underlying base shape is too narrow 
to sample sufficiently large and diverse patches, resulting in poor synthesis results with artifacts and irregular patterns. Conversely, when multiple trusses are placed together to construct a larger roof (first row), the underlying base shape becomes large enough to capture patches that maintain the structural patterns, resulting in more satisfactory synthesized textures. 
Additionally, the perforated structure on the Lego loader shown in Fig.~\ref{fig:limitation} is too narrow to sample any patches, posing a significant challenge for subsequent synthesis. On the other hand, our method is capable of easily extracting and synthesizing the bump texture on the Lego base.

\begin{figure}[H]
	\centering
	\includegraphics[width=1.\linewidth]{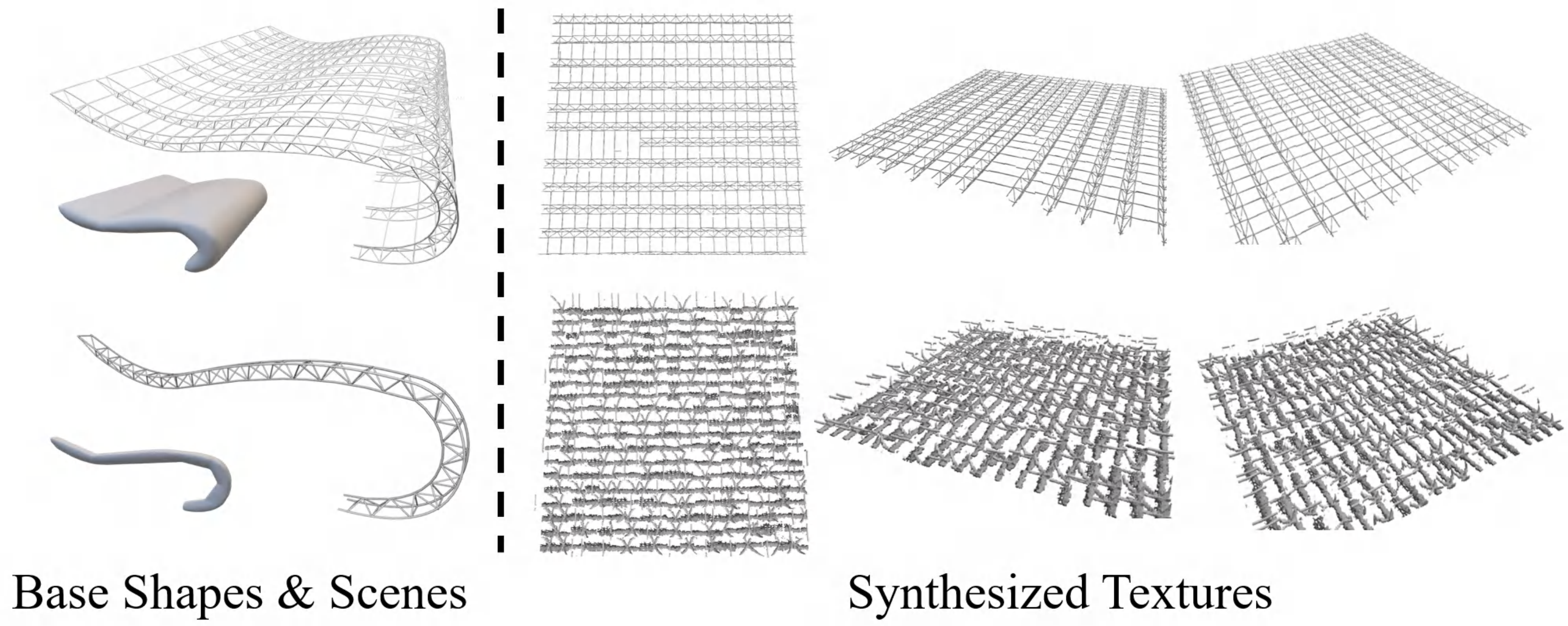}
	\vspace{-2mm}
	\caption{\textbf{Challenges in Limited Coarse Extents.} The repeated structures of the truss pattern are distributed on a long but narrow coarse surface, which makes sampled patches too small to synthesize high-quality textures. The roof composed of trusses, on the contrary, has enough coarse extents to capture and synthesize satisfactory results.}
\label{fig:limitation_capture}
\end{figure}

\begin{figure*}[thbp]
	\centering
	\includegraphics[width=0.95\linewidth]{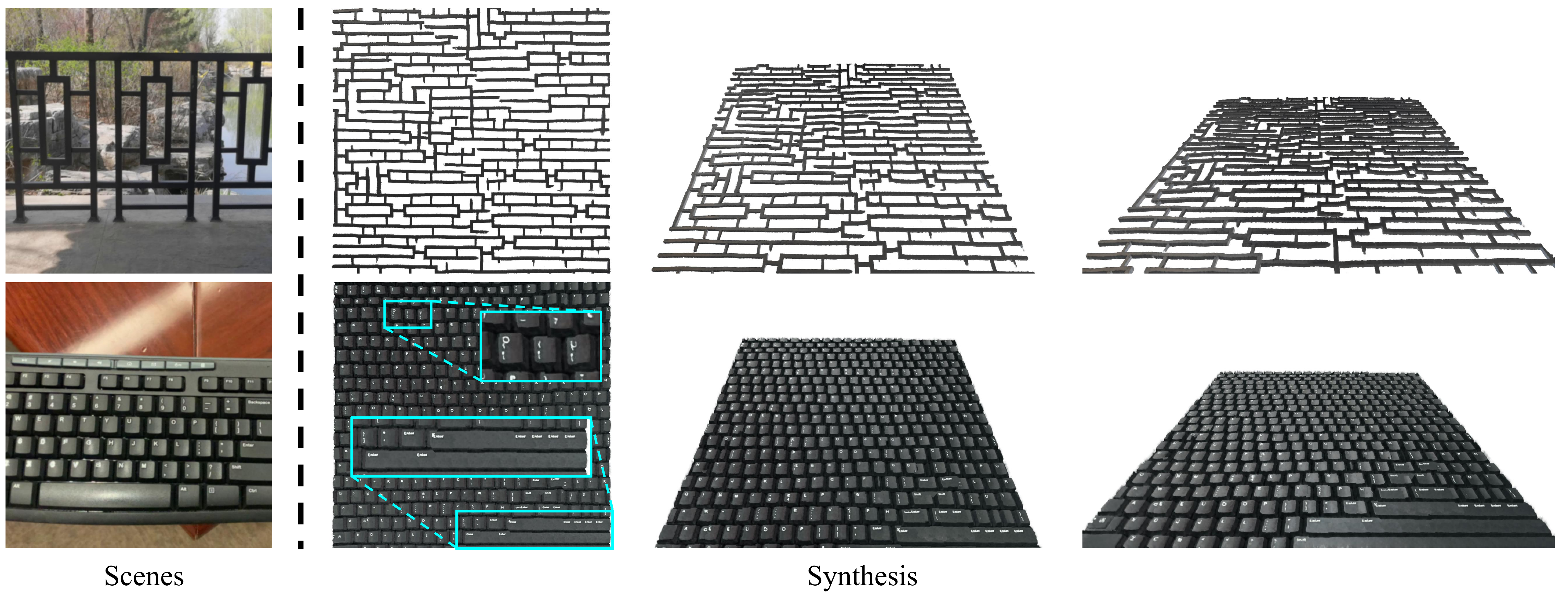}
	\vspace{-2mm}
	\caption{\textbf{Challenges on Texture Synthesis.} Our synthesis approach based on patch matching struggles to exactly preserve the continuity of highly structured textures (first row) requiring strict matching with a few captured exemplars. Our method is also agnostic to the semantic contents of textures like the keycap texture of a keyboard (second row).}
\label{fig:limitation2}
\end{figure*}

\noindent\textbf{Complex Topology} For scenes with complex topology, such as the Lego example shown in Fig.~\ref{fig:limitation}, accurately representing the detailed surface with a coarse mesh becomes challenging. Consequently, capturing the desired bump texture on the inner surface of the Lego cockpit becomes difficult. Additionally, the rendering quality of NeRF-Texture may degrade in such scenarios, leading to artifacts as seen in the zoomed-in section of Fig.~\ref{fig:limitation}. This degradation is due to the coarse shape approximation, which fails to provide an optimal parameterization for NeRF.

\subsection{Challenging Texture Synthesis}
The challenges in texture synthesis arise when textures \textbf{1)} require strict matching or \textbf{2)} have semantic contents.

\noindent\textbf{Strict Matching} Our patch-matching approach selects patches from the best-matched candidates based on the matching errors between patches and synthesized regions. However, this approach is a local optimal and not necessarily a global one since the selected candidate may cause significant matching errors in subsequent iterations. Therefore, it is a greedy strategy and may cause the breaking of structures for textures requiring strict matching, especially when only limited patch exemplars are available. As shown in the first row of Fig.~\ref{fig:limitation2}, although the overall structure of the synthesized railing texture is preserved, there are still some breaking parts in the connection of rails.

\noindent\textbf{Semantic Content} Our synthesis algorithm is semantically agnostic, which can distort semantic content such as keycap shapes and incorrect synthesis of characters, as shown in the second row of Fig.~\ref{fig:limitation2}. The letter P and bracket symbol are printed on the same keycap in the zoom-in window. The space key is extended to an unreasonable length. To address this limitation, our approach could potentially incorporate recent generative techniques, such as Diffusion Models~\cite{ho2020denoising}.

}

\section{Conclusion}
\label{sec:conclusion}
We present NeRF-Texture, a novel approach that captures, models, synthesizes, and renders real-world textures with rich geometric and appearance details.
A coarse-fine decomposition representation is introduced to disentangle the meso-structure texture and base shape.
Based on the representation, we adopt a latent patch-matching algorithm to synthesize acquired textures on the UV plane or arbitrary surfaces.
A clustering constraint regularizes the distribution of latent features for better synthesis.

\section{Acknowledgement}

This work was supported by Beijing Municipal Natural Science Foundation for Distinguished Young Scholars (No. JQ21013), National Natural Science Foundation of China (No. 62322210) and Beijing Municipal Science and Technology Commission (No. Z231100005923031).



\ifCLASSOPTIONcaptionsoff
  \newpage
\fi

{\small
    \bibliographystyle{IEEEtran}
    \bibliography{egbib}
}

\end{document}